\documentclass{style/ar-1col}
\usepackage[numbers]{natbib}
\usepackage{url}
\usepackage{soul}
\usepackage{amssymb}
\usepackage{amsfonts}
\usepackage{mathrsfs}
\usepackage{array}
\usepackage{amsmath}
\newcommand{\argmin}{\operatornamewithlimits{argmin}}

\usepackage{nameref}

\usepackage{algorithm}
\usepackage{algpseudocode}

\newcommand{\R}{\mathbb{R}}
\newcommand{\HybridControlSystem}{\mathscr{HC}}
\newcommand{\HybridSystem}{\mathscr{H}}

\newcommand{\Domain}{\mathcal{D}}
\newcommand{\Guard}{S}

\newcommand{\ControlInput}{\mathcal{U}}
\newcommand{\ResetMap}{\Delta}

\newcommand{\DirectedGraph}{\Gamma}
\newcommand{\Vertex}{V}
\newcommand{\Edge}{E}
\newcommand{\vi}{{v}}
\newcommand{\ei}{{e}}

\newcommand{\Xext}{\mathbb{X}_{\mathrm{ext}}}

\newcommand{\secref}[1]{Section \ref{#1}}
\newcommand{\figref}[1]{Figure \ref{#1}}
\newcommand{\eqnref}[1]{Equation \ref{#1}}

\jname{Annu. Rev. Control Robot. Auton. Syst}
\jvol{}
\jyear{2021}
\doi{10.1146/DOI}

\begin{document}

\markboth{Reher and Ames}{Dynamic Walking}

\title{Dynamic Walking: \\
Toward Agile and Efficient Bipedal Robots
}

\author{Jenna Reher$^1$ and Aaron D. Ames$^2$
\affil{$^1$Mechanical and Civil Engineering, California Institute of Technology, Pasadena CA, USA, 91125; email: jreher@caltech.edu}
\affil{$^2$Mechanical and Civil Engineering, Control and Dynamical Systems, California Institute of Technology, Pasadena CA, USA, 91125; email: ames@caltech.edu}}

\begin{abstract}
Dynamic walking on bipedal robots has evolved from an idea in science fiction to a practical reality.  This is due to continued progress in three key areas: a mathematical understanding of locomotion, the computational ability to encode this mathematics through optimization, and the hardware capable of realizing this understanding in practice.  In this context, this review article outlines the end-to-end process of methods which have proven effective in the literature for achieving dynamic walking on bipedal robots.   We begin by introducing mathematical models of locomotion, from reduced order models that capture essential walking behaviors to hybrid dynamical systems that encode the full order continuous dynamics along with discrete footstrike dynamics.  These models form the basis for gait generation via (nonlinear) optimization problems.  Finally, models and their generated gaits merge in the context of real-time control, wherein walking behaviors are translated to hardware.  The concepts presented are illustrated throughout in simulation, and experimental instantiation on multiple walking platforms are highlighted to demonstrate the ability to realize dynamic walking on bipedal robots that is agile and efficient. 
\end{abstract}

\maketitle

\section{INTRODUCTION} \label{sec:intro}

The realization of human-like capabilities on artificial machines has captured the imagination of humanity for centuries. The earliest attempts to realize this were through purely mechanical means. In $1495$, Leonardo da Vinci detailed his \textit{Automa cavaliere}, a primitive humanoid in a knights armor and operated by a number of pulleys and cables. However, these mechanical automatons lacked the ability to apply feedback control and thus the field remained largely dormant until digital computers became broadly available.  In $1921$ the word “robot” was coined by Czech playwriter Karel Carek, just $40$ years before microprocessors were introduced and soon thereafter the field of legged robots began to emerge.  

Today, the field of robotic legged locomotion is of special interest to researchers as humans increasingly look to augment their natural environments with intelligent machines. In order for these robots to navigate the unstructured environments of the world and perform tasks, they must have the capability to reliably and efficiently locomote. The first control paradigms for robotic walking used a notion of \textit{static stability} where the vertical projection of the Center of Mass (COM) is contained to the support polygon of the feet, leading to the WABOT 1 robot in the early $1970$s at Waseda university \cite{lim2007biped} and the first active exoskeletons by Vukobratovi\'c at the Mihailo Puppin Institute \cite{vukobratovic1974development}. This static stability criterion was very restrictive, leading to the development of the \textit{Zero Moment Point} criterion \cite{vukobratovic1972stability, vukobratovic2004zero}, which enabled a wider range of robotic locomotion capabilities by generalizing from the COM to the Center of Pressure (COP). Despite this generalization, it still restricts the motion of the robot to be relatively conservative and does not allow for more dynamic motions when compared to the capabilities of biological walkers \cite{collins2005efficient}. Nevertheless, this methodology has been perhaps the most popular methodology to date for realizing robotic locomotion. 
Several of this method has been applied to various successful humanoid robots such as the Honda ASIMO robot 
\cite{hirose2007honda}, the HRP series \cite{KaKaKaYoAkKaOtIs2002,AkKaKaOtMiHiKaKa2005,ItNoUrNaKoNaOkIn2014}, and HUBO \cite{PaKiLeOh2005}.

As the field progressed into the 1980's it became clear that to achieve truly dynamic locomotion it was necessary to further exploit the natural nonlinear dynamics of these systems in an energy efficient and stable fashion.  In stark contrast to the concept of fully actuated humanoid locomotion, Mark Raibert and the LegLab launched a series of hopping robots which demonstrated running behaviors and flips \cite{raibert1984experiments,raibert1986legged}. 
To achieve these behaviors, there was a shift from the conservative walking models encoded by the zero moment point to reduced order models (e.g. the spring loaded inverted pendulum) that ensure dynamic locomotion through the creation of stable periodic orbits \cite{hurmuzlu1986role}. Building upon this core idea, Tad McGeer began development of completely passive walking machines, which would ultimately give rise to the field of \textit{passive dynamic walking} \cite{mcgeer1990passive}. The downside of this method is that the system has little to no actuation with which it can respond to perturbations or to perform other tasks. However, these breakthroughs were critical in demonstrating that dynamic robotic locomotion was possible on systems which were not fully actuated, and that this underactuation could actually be leveraged to improve their performance.

Despite the advances leading up to the turn of the century, there remained a growing gap between the physical capabilities of robotic systems and the development of controllers to exploit them. 
This was particularly stark in the area of underactuated walking, where
the lack of formal approaches that leverage the intrinsically nonlinear dynamics of locomotion limited the ability to fully exploit the robot's actuation authority.   
In the early $2000$'s, a key contribution to this area was introduced by 
Jessy Grizzle et al. \cite{westervelt2003hybrid} in which they developed the notion of \textit{virtual constraints}, or holonomic constraints enforced via control rather than a 
physical mechanism. Enforcing these constraints leads to low-dimensional invariant surfaces, the \emph{zero dynamics surface}, in the continuous phase of the model. These virtual constraints could then be designed such that this surface is \textit{hybrid invariant} - being invariant under both the continuous and discrete dynamics - ultimately leading to the concept of \textit{Hybrid Zero Dynamics} (HZD) \cite{westervelt2018feedback}. The end result is formal guarantees on the generation and stabilization of periodic orbits \cite{ames2014rapidly}, i.e., walking gaits. This paradigm for control of dynamic underactuated locomotion has pushed boundaries on what is achievable, including: fast running
\cite{sreenath2013embedding, ma2017bipedal} and efficient humanoid walking \cite{reher2016durusrealizing,reher2016durusmulticontact}. 

\begin{figure}[b!]
	\centering
	\includegraphics[width=.88\columnwidth]{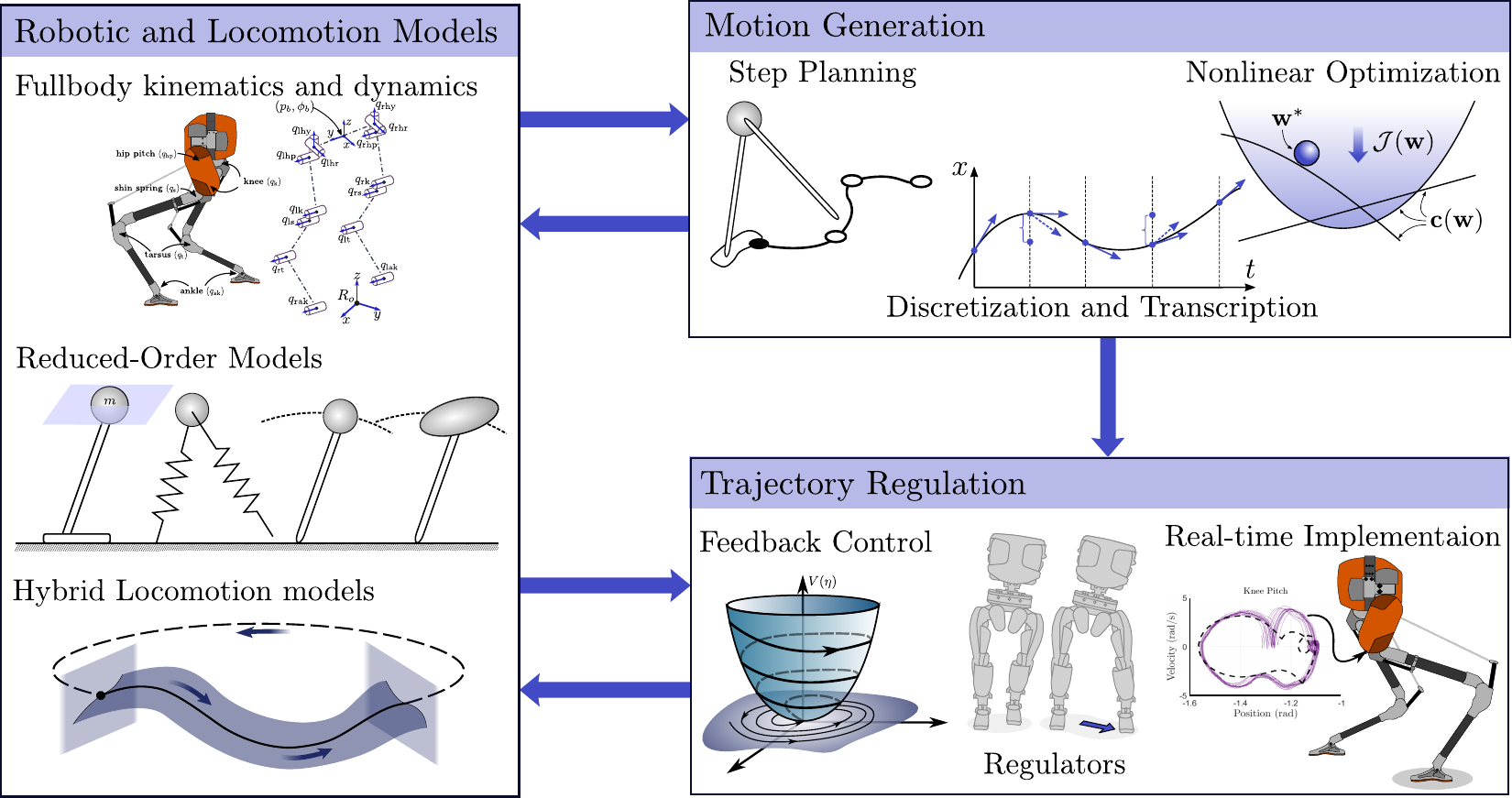}
	\caption{Dynamic walking is a complex behavior, requiring control designers and roboticists to simultaneously consider: robotic models, the transcription of locomotion into a motion planning problem, and the coordination and actuation of the system via control laws. Depicted here is the interconnection of these components, which provides an outline for this review article.}
	\label{fig:meta-algorithm}
\end{figure}

As we examine this brief historical outline of key developments in dynamic walking, it can be observed that with each new proposed methodology comes a greater understanding of how to model, plan, and execute increasingly complex behaviors on these robotic systems. Due to the inherently difficult nature of dynamic walking, successes in the field have typically been achieved by considering all aspects of the problem, often with explicit consideration of the interplay between modeling and feedback control (see Fig. \ref{fig:meta-algorithm}).
Specifically, the robotic and locomotive models which are used ultimately inform the planning problem and therefore the resulting behavior. Controllers which can actuate and coordinate the limbs must then be developed which, ideally, provide tracking, convergence and stability guarantees. In this review, we therefore examine how this interplay between modeling, motion planning, and trajectory regulation has shaped the dynamic walking on bipedal robots to date.  

The remainder of this paper is structured as follows. In Section \ref{sec:models} we present reduced-order models that have provided canonical examples of dynamic locomotion, and in Section \ref{sec:hybrid} we introduce extensions to hybrid system models for dynamic walking.
Section \ref{sec:motion_planning} discusses how these models have been used to generate stable walking motions through various motion planning approaches and corresponding optimization problems. Finally, Section \ref{sec:regulation} provides several existing methods for controlling the robot as informed by the methods introduced in the earlier sections, with a view toward hardware realization. 
This interconnection can be seen in Figure \ref{fig:meta-algorithm}, where each subsequent section is informed by the prior. 

\section{DYNAMIC MODELS OF BIPEDAL LOCOMOTION} \label{sec:models}

\begin{figure}[t!]
	\centering
	\includegraphics[width=1\columnwidth]{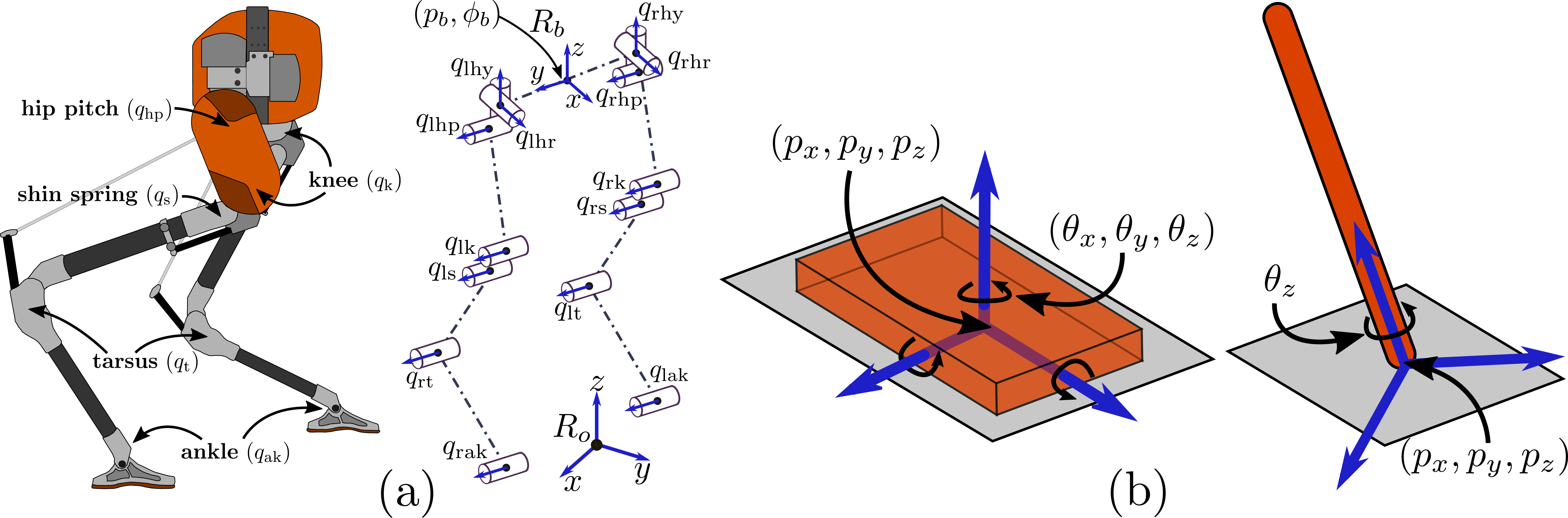}
	\caption{A visual demonstration of the robotic configuration and contact constraints which are applied to the robot. (a) The floating-base coordinate system for a Cassie bipedal robot, with a coordinate frame attached to the hip and rotational joints connecting rigid linkages of the body. (b) Contact geometry of the constraints for an underactuated flat-foot contact and a point-foot contact.}
	\label{fig:dynamic_model_contact_combined}
\end{figure}

In this section, we provide background on the modeling of dynamic bipedal robots and contextualize several of the most popular approaches for encoding or approximating locomotion via reduced order models. 
A unifying theme among the broad spectrum of models used for legged locomotion, both in this section and the next,
is that the system must undergo intermittent contact with the surrounding environment in order to move. This fact is inexplicably tied to legged locomotion. How the overall walking system is ultimately modeled plays a critical role in the planning and control approaches that realize locomotion.

\subsection{Bipedal Robots: Floating Base Systems with Contacts} \label{sec:fullbody_model}
Bipedal robotic platforms are conveniently modeled using a tree-like structure with an ordered collection of rigid linkages. This structure lends itself well to generalization, and thus tools to facilitate the generation of symbolic \cite{hereid2017frost} or algebraic \cite{featherstone2014rigid} expressions for the kinematics and dynamics of the robot are commonly used. The robot itself must ambulate through a sequence of contact conditions with the environment. Because interactions with the environment are always changing, a convenient method for modeling the system is to construct a representation of the robot in a general position, and then enforcing ground contacts through forces arising from the associated holonomic constraints that are imposed at the feet. This is often referred to as the \emph{floating-base model} of the robot.

\subsubsection{The Configuration Space} 
In order to represent the floating-base, let $R_0$ be a fixed inertial frame attached to the world and let $R_b$ be a body reference frame rigidly attached to the pelvis of the robot with the origin located at the center of the hip. Then the Cartesian position $p_b\in\mathbb{R}^3$ and orientation $\psi_b \in SO(3)$ compose the floating base coordinates of frame $R_b$ with respect to $R_0$. The remaining coordinates which dictate the shape of the actual robot, $q_l\in\mathcal{Q}_l \subset \R^{n_l}$, are the local coordinates representing rotational joint angles and prismatic joint displacements.
An image of this floating base coordinate system definition for a Cassie bipedal robot is given in \figref{fig:dynamic_model_contact_combined}(a). 
The combined set of coordinates is $q = (p_b^T, \phi_b^T, q_l)^T \in \mathcal{Q} = \mathbb{R}^3 \times SO(3) \times \mathcal{Q}_l$ with the states $x = (q^T,\dot{q}^T)^T \in T\mathcal{Q} = X$.

\subsubsection{Continuous Dynamics}
Traditional methods for modeling the dynamics of floating-base systems typically result in the separation of the equations of motion into multiple parts \cite{pfeiffer1996multibody}; one arising from the multibody continuous dynamics, and the other imposed via constraints on contacts with the environment. If we continue with the assumption that the robot structure is a rigid collection of linkages then we can consider the continuous dynamics of a bipedal robot in the Lagrangian form (see \cite{murray1994mathematical}):
\begin{align}
    D(q)\ddot{q} + H(q, \dot{q}) &= Bu + J_h(q)^T \lambda, \label{eq:eom_continuous}
\end{align}
where $D(q)$ is the inertia matrix, $H(q, \dot{q})$ contains the Coriolis and gravity terms, $B$ is the actuation matrix, $u \in U \subseteq \mathbb{R}^m$ is the control input, and the Jacobian of the holonomic constraints applied to the robot is $J_h(q)=\frac{\partial h}{\partial q}(q)$ with the corresponding wrenches $\lambda\in\R^{m_{h}}$. 
These dynamics can be expressed in a state-space representation as:
\begin{align}
    \frac{d}{dt} \begin{bmatrix} q \\ \dot{q} \end{bmatrix} 
    = 
    \underbrace{\begin{bmatrix} \dot{q} \\ D^{-1}(q) \left( J_h(q)^T \lambda - H(q,\dot{q}) \right) \end{bmatrix}}_{f(x)}
    +
    \underbrace{\begin{bmatrix} 0 \\ D(q)^{-1} B \end{bmatrix}}_{g(x)} u. \label{eq:manip_nl_eom}
\end{align}

\subsubsection{Contact Forces}
The fact that the robotic model is derived using a floating-base representation means that as we manipulate the robot, the resulting ground force interaction through the Lagrangian dynamics in \eqnref{eq:eom_continuous} is critical. The most popular method for modeling ground interaction is to assume rigid contacts with nonpenetration, the resulting forces are then considered to be \textit{unilateral} \cite{pfeiffer1996multibody}, meaning that they can push and not pull on the ground. The resulting normal force cannot be negative, $\lambda_z \geq 0$, and this implies that when this condition crosses zero that the contact will leave the ground. A point of the robot in static contact with the ground will satisfy a closure equation of the form:
\begin{equation}
    \eta(q) = \begin{bmatrix} p_c(q)^T, & ~ \phi_c(q)^T \end{bmatrix}^T  = \mathrm{constant},
    \label{eq:contact_point}
\end{equation}
where $p_c(q)$ is the Cartesian position of the contact point and $\phi_c(q)$ is a rotation between contacting bodies \cite{grizzle2014models}. 
Differentiating twice yields acceleration constraints on the robot:
\begin{align}
    J_h(q) \ddot{q} + \dot{J}_h(q, \dot{q}) \dot{q} &= 0, \label{eq:Jeom}
\end{align}
leading to a system of equations, with Equations \ref{eq:eom_continuous} and \ref{eq:Jeom} coupling the accelerations to the inputs and resulting constraint forces. The geometry of robotic feet is often given as either a flat foot or a single point of contact, shown in \figref{fig:dynamic_model_contact_combined}(b). 
Assuming three non-collinear points of contact, the foot can be modeled as a flat plane. The position and orientation of the plane with respect to the ground will then create a $6$ degree-of-freedom (DOF) closure constraint in \eqnref{eq:contact_point} ($m_h = 6$). Alternatively, many underactuated robots have point-feet. If the assumption is made that the foot will not yaw while in contact, then this will form a $4$ DOF constraint on the Cartesian positions and rotation about the $z$-axis ($m_h = 4$). 

Finally, when designing motions for the robot it is important to also model the real-world limitations to the allowed tangential force before it will break a nonslip condition. The most popular approach is to employ a classical \textit{Amontons-Coulomb model} of (dry) friction \cite{grizzle2014models}. For a friction coefficient $\mu$, the space of valid reaction forces is characterized by the \textit{friction cone}:
\begin{align}
    \mathcal{C} = \left\{ \left. ( \lambda_x, \lambda_y, \lambda_z ) \in \mathbb{R}^3 \right| \lambda_z \geq 0; \sqrt{\lambda_x^2 + \lambda_y^2} \leq \mu \lambda_z \right\}. \label{eq:cone_friction}
\end{align}

\subsection{Linear Inverted Pendulum and the Zero Moment Point} \label{sec:zmp}


\begin{figure*}[t]
\centering
	\includegraphics[width=1\columnwidth]{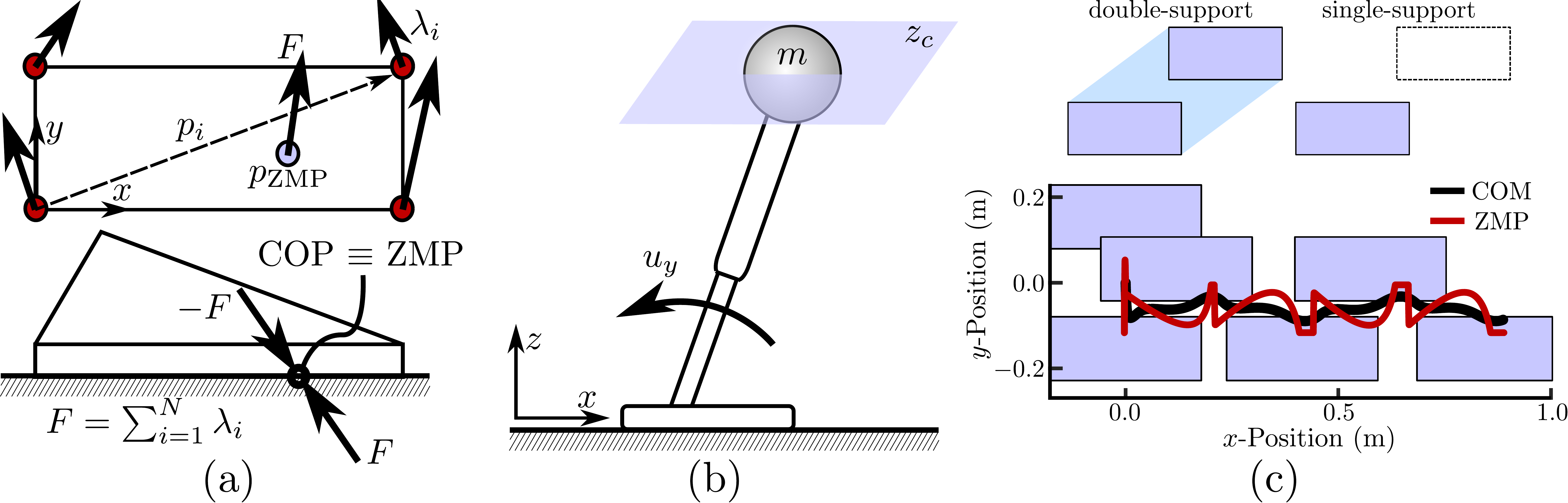}
	\caption{A depiction of the principles and modeling assumptions of the LIPM approach: (a) Visualization of the of the ZMP, where the foot is ``dynamically balanced'' if the resultant force $F$ is within the support polygon. (b) LIPM with a telescoping leg and actuated ankle to control the robot along a horizontal surface. (c) Support polygon and an example of a  planned ZMP trajectory.}
	\label{fig:zmp_lip_example}
\end{figure*}

In this section, we describe the basic aspects of the \textit{Zero Moment Point} (ZMP) and how it has been used in linear inverted pendulum models (LIPM) of locomotion. The concept of the ZMP is identical to the center of pressure (COP), and was originally introduced through a series of observations on the stability of anthropomorphic walkers by Vukobrativi\`c in the early $1970$'s \cite{vukobratovic1970stability, vukobratovic1972stability}. The primary interpretation of the ZMP is: \textit{the point on the ground at which the reaction forces between the robot's contacts and the ground produce no horizontal moment}. Consider a robot standing in single-support, with a finite number of contact points ($p_i$) that constrain the foot to be flat. As shown in \figref{fig:zmp_lip_example}(a), the resultant forces will consist of normal ($\lambda_n$) and tangential components ($\lambda_t$). The ZMP is then computed as:
\begin{align}
    p_{\mathrm{ZMP}} := \frac{\sum_{i=1}^{N} p_i \lambda_{i,n}}{\sum_{i=1}^{N} \lambda_{i,n}}. \label{eq:zmp_point}
\end{align}
This led to perhaps the most commonly used dynamic stability margin \cite{arakawa1997natural, shih1990trajectory, yamaguchi1993development, sardain2004forces, hirukawa2006universal}, referred to as the \textit{ZMP criterion}, which states that a movement is stable so long as the ZMP remains within the convex hull of the contact points (also known as the \emph{support polygon}). This notion is conservative, and controlling these motions typically require the robot to remain fully actuated, with position controlled joints and load cells in the feet. 

The ZMP criterion has been tied extensively to the linear inverted pendulum model (LIPM) in order to considerably simplify the trajectory design process, as the ZMP can be written explicitly in terms of the COM dynamics \cite{kajita2001real}. This has led to many researchers to consider a Newton-Euler representation of the \emph{centroidal dynamics}, written as: 
\begin{align}
    m (\ddot{c} + g) = \sum_i \lambda_i , ~ ~ ~ 
    \dot{L} = \sum_i(p_i-c) \times \lambda_i,
    \label{eq:centroid_dynamics}
\end{align}
with $c$ the COM position, $L = \sum_k (x_k-c) \times m_k \dot{x}_k + \mathbf{I}_k \omega_k$ the angular momentum, $g$ gravitational acceleration,  $\lambda_i$ the contact forces, $p_i$ is each contact force position, $\dot{x}_k$, $\omega_k$ the linear and angular velocities on the $k$-th linkage, $m_k$, $\mathbf{I}_k$ are the masses and inertia tensors, and $m$ the total mass of the robot. 
If we constrain the motion of a fully actuated inverted pendulum with a massless telescoping leg such that the COM moves along a horizontal (x,y) plane, then we obtain a 
linear expression for the 
robot dynamics. 
An example of the LIPM is visualized in \figref{fig:zmp_lip_example}(b). The dynamics of the LIPM at a given height,
$z_c$, is: 
\begin{align}
    \ddot{x} = \frac{g}{z_c}x + \frac{1}{m z_c} u_y, \qquad \ddot{y} = \frac{g}{z_c}y + \frac{1}{m z_c} u_x, 
\end{align}
where $m$ is the mass of the robot, $g$ is the acceleration of gravity, and $u_x$, $u_y$ are the torques about the $x$ and $y$ axes of the attachment to the ground, i.e., the ankle. The ZMP location on the ground can also be directly written in terms of the LIPM dynamics as:
\begin{align}
    p_{\mathrm{ZMP}}^x = x - \frac{z_c}{g} \ddot{x}, \qquad p_{\mathrm{ZMP}}^y = y - \frac{z_c}{g} \ddot{y}.
\end{align}
The LIPM can be viewed as a cart-table system \cite{vukobratovic2004zero}, where the cart-table lies on a base with a geometry corresponding the the support polygon \cite{kajita2003biped}. 
Designing walking with the ZMP can be essentially reduced to an inverse kinematics problem, where the primary planning is done on the ZMP trajectory. 
\figref{fig:zmp_lip_example}(c) shows an example ZMP trajectory for several forward steps, wherein the trajectory for this walking is planned so that the ZMP always stays within the support polygon. 
ZMP walking has largely been applied to humanoids, 
such as 
the WABIAN robots \cite{lim2007biped}, HRP series \cite{KaKaKaYoAkKaOtIs2002,ItNoUrNaKoNaOkIn2014}, Johnnie \cite{pfeiffer2002concept}, and HUBO \cite{PaKiLeOh2005}. 


\subsection{Capturability and Nonlinear Inverted Pendulum Models} \label{sec:nlpend_capturepoint}

\begin{figure*}[b]
\centering
	\includegraphics[width=0.95\columnwidth]{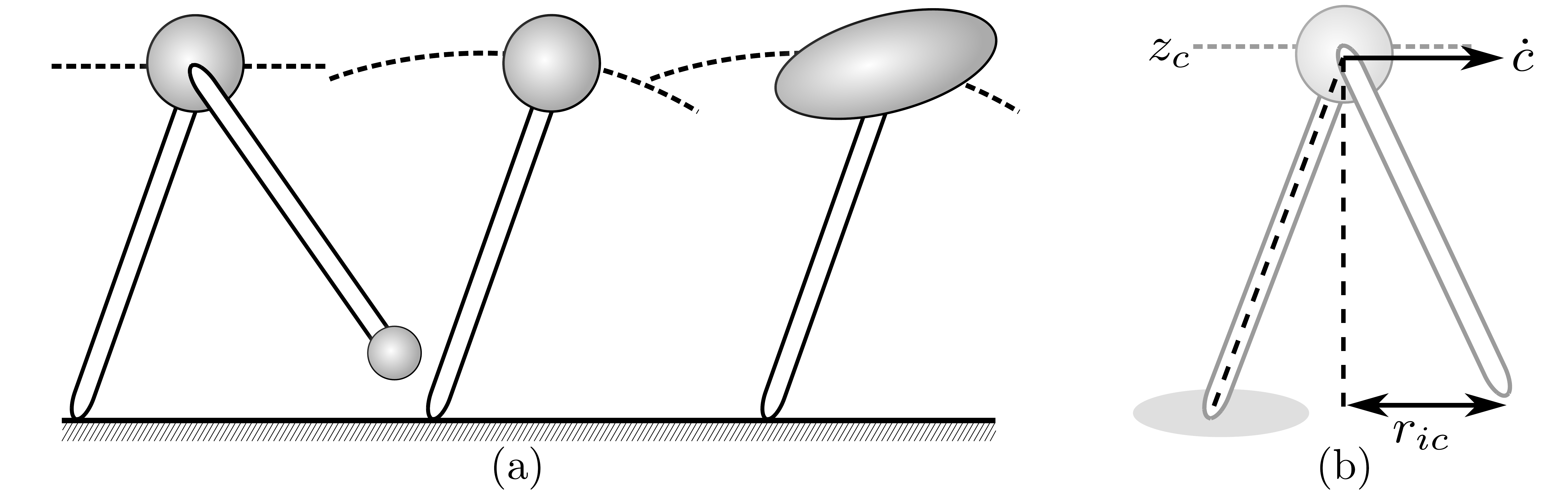}
	\caption{(a) A depiction of several variations on inverted pendulum models, which attempt to expand the possible behaviors of the robot by accounting for more of the body inertia or by releasing the constrained motion of the hip. (b) A depiction of the capture point for a LIP walking robot.}
	\label{fig:nl_pendulum_capturept}
\end{figure*}

Rather than characterize the stability of walking based on the ZMP, Pratt \cite{pratt2006capture} and Hof \cite{hof2008extrapolated} independently introduced the idea of a \textit{Capture Point} (CP), referred to as the ``extrapolated center of mass'' (XCOM) by Hof. The CP can be intuitively described as the point on the ground onto which the robot has to step to come to a complete rest. 
In canonical examples of the CP methods, the overall walking motions of the robot are planned and controlled based on the \textit{(instantaneous) capture point} (ICP) dynamics. In this case, the COM of the robot is constrained to move at a constant height along a horizontal plane, and thus uses a LIP representation of the robotic system. 
It was shown in \cite{koolen2012capturability} that for the compound variable $r_{ic}^{x,y} = c + \sqrt{\frac{z_c}{g_z}}\dot{c}$, that the unstable portion of the resulting system dynamics (along the horizontal direction) can be written in a constrained fashion as follows:
\begin{align}
    \dot{r}_{ic}^{x,y} = \sqrt{\frac{g_z}{z_c}} (r_{ic}^{x,y} - r_{CMP}^{x,y}),
    \qquad \textrm{subject to:} \qquad
    r_{ic}^{x,y} \in \mathrm{conv}\{ p_i^{x,y} \},
    \label{eq:icp_dynamics}
\end{align}
where $r_{ic}^{x,y}$ is the horizontal location of the ICP. The main consideration of the locomotion process is then to ensure that feet are placed such that $r_{ic}^{x,y}$ lies within the support polygon. 
Satisfying this condition means that the COM will converge to the CP and come to a rest. Despite this intuitive representation of stability, the LIPM walking simplifications come with a steep cost due to the stringent requirements on the motion and actuation of the robot. 
Yet it is precisely these characteristics which make the model most suitable for performing complex multi-objective tasks which include manipulation during intermittent conservative motions. The maturity and reliability of the LIPM made it prevalent in the walking controllers used at the DARPA Robotics Challenge \cite{johnson2015team, kuindersma2016optimization,dedonato2015human}.

%
In an attempt to overcome 
issues associated with the strict assumptions of the LIPM, researchers have introduced 
more complex pendulum models 
(illustrated in Figure \ref{fig:nl_pendulum_capturept}(a)).
The largest constraint on 
LIPM walking is the constant center of mass height assumption, leading to the development of a nonlinear inverted pendulum with variable mass height \cite{pratt2007derivation}. To 
account for the inertia of a swinging leg, the addition of a mass at the swing foot was proposed and termed the Gravity Compensated LIPM \cite{park1998biped}. 
One of the most commonly used models 
in
the literature to address nontrivial angular momenta from the limbs of large robots moving dynamically is to add a flywheel to the 
hip, 
which can be used to represent the 
inertia of the robot body as it moves \cite{stephens2011push}. 
A flywheel model of the robot has gained recognition as a convenient representation of the robotic system particularly for CP control \cite{pratt2006capture,pratt2012capturability}. 
Various pendulum models have been widely used in analysis of push recovery and balance \cite{hof2005condition,hyon2007full,stephens2007humanoid,takanishi1990control}. 
The CP approach has also been used to demonstrate walking 
successfully on hardware \cite{englsberger2011bipedal,koolen2012capturability}, and was famously used on Honda's ASIMO 
\cite{takenaka2009realpaper1,KaTsTs04}.

\subsection{Spring Loaded Inverted Pendulum} \label{sec:slip}

\begin{figure*}[b]
\centering
	\includegraphics[width=1\columnwidth]{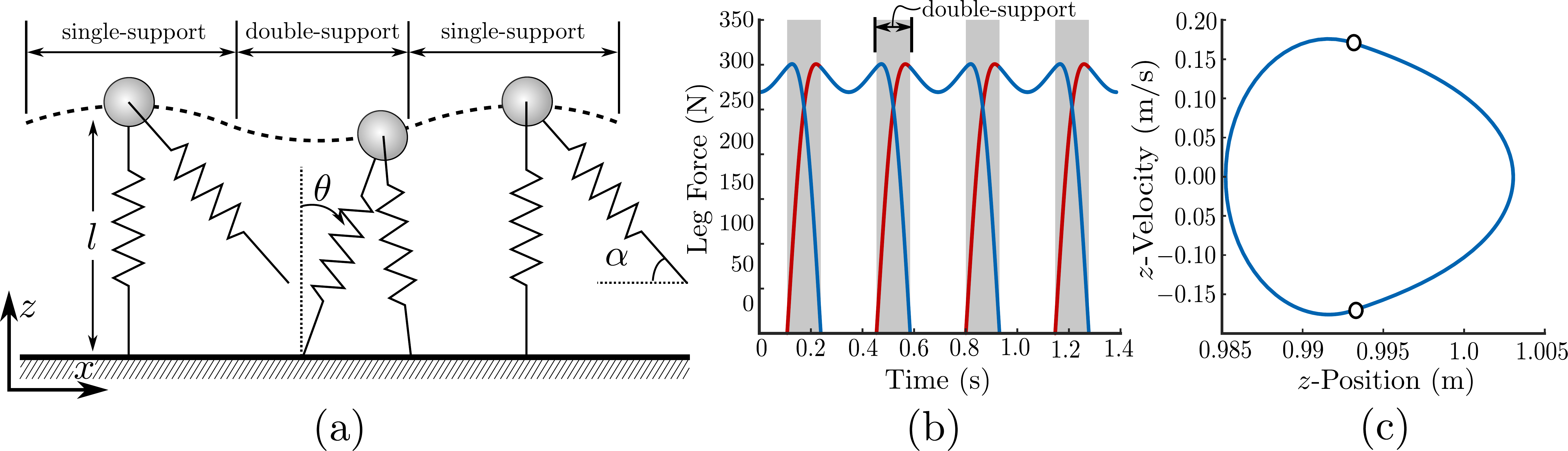}
	\caption{
	(a) The SLIP model, with the mass 
	at the hip and virtual compliant legs. (b) 
	Contact forces 
	during walking, where the ``double hump'' profile is observed 
	in biological walkers. 
	(c) A periodic orbit for the vertical COM 
	coordinate, where a lack of impact yield no footstrike discontinuity.}
	\label{fig:slip_figure_bar}
\end{figure*}

Classic work by Raibert on hopping and running robots in the $1980$s demonstrated the efficacy of using compliant models in locomotion through the development of a planar hopper which could bound at a speed of $1$ m/s \cite{raibert1984experiments} and a 3D hopper which could achieve running without a planarizing boom \cite{raibert1986legged}. These early successes drove researchers to investigate a \emph{Spring Loaded Inverted Pendulum} (SLIP) representation of bipedal robots, shown in \figref{fig:slip_figure_bar}(a). The SLIP model provides a low-dimensional representation of locomotion which draws inspiration from biological studies on animal locomotion \cite{geyer2006compliant, blickhan1989spring}. The SLIP is particularly attractive due to its inherent efficiency and robustness to ground height variations.

In order to use this model to synthesize controllers for actual robots, the control objectives are typically decomposed into three components: (1) achieving a particular footstrike location to regulate forward speed, (2) injecting energy either through passive compliance or motors to regulate the vertical height of the CoM, and (3) regulating the posture of the robot. One then designs the walking and running motions with SLIP models and compensates for model mismatch or disturbances with well tuned foot placement style controllers \cite{pratt2001virtual, ahmadi2006controlled, altendorfer2004stability, rummel2010stable, zeglin1998control} (see \secref{sec:footplacement}). 
To this end, the dynamics of the SLIP are derived by assuming that the mass of the robot is concentrated at the hip with virtual springy legs:
\begin{align}
	0 &= m \ddot{l} - m l \dot{\theta}^2 + m g \sin(\theta) + F_{\mathrm{slip}} \label{eq:slip-dynamics} \\
	0 &= m [l^2 \ddot{\theta} + 2 l \dot{l} \dot{\theta}] + m g l \sin(\theta), \nonumber
\end{align}
where $l$ is the stance leg length, $\theta$ is the stance leg angle, and $F_{\mathrm{slip}}$ is the force arising from the spring compression. One of the signature characteristics of this model is the ``double hump'' profile of the reaction forces, shown in \figref{fig:slip_figure_bar}(b), described by the force interactions observed in biological walkers \cite{blickhan1989spring}. A key contribution introduced by the SLIP community is the handling of underactuated behaviors, with many of the corresponding robots having point-feet and flight phases of motion. Finding a stable gait thereon does not rely on the quasi-static assumptions used for the fully actuated pendulum walkers of the preceding sections---instead focusing on stable cyclic locomotion.
Dynamic stability is defined based on a constraint on the periodicity of the walking (detailed in the sidebar ``Periodic Notions of Stability'').  To achieve forward walking, the initial states of the robot and the angle of attack $\alpha$ for the swing leg are chosen to yield a periodic gait; see \figref{fig:slip_figure_bar}(c). It is important to note that since the legs are massless, impacts are not considered, and the resulting orbit will be closed with no instantaneous jumps in the velocity.

\begin{textbox}[t]
\section{Periodic Notions of Stability} 
\label{sidebar:periodic}
One can view steady state walking as a periodic motion which is not instantaneously stable, but is stable from footstrike to footstrike \cite{hobbelen2007limit}---in other words, walking is ``controlled falling.'' 
Concretely, a solution $\phi(t,t_0,x_0)$ to a dynamical system $\dot{x} = f(x)$ is \textit{periodic} if there exists a finite $T>0$ such that $\phi(t+T, t_0, x_0)=\phi(t,t_0,x_0)$ for all $t \in [ t_0, \infty)$ and a set $\mathcal{O} \subset X$ is a \textit{periodic orbit} if $\mathcal{O}= \{ \phi(t,t_0,x_0) | t \geq t_0 \}$ for some periodic solution $\phi(t,t_0,x_0)$. In a seminal paper on passive dynamic walking \cite{mcgeer1990passive}, McGeer popularized the method of Poincar\`e to determine the existence and stability of periodic orbits \cite{perko2013differential} for walking. In this approach, one step is considered to be a mapping from the walkers state $x_k$ at a definite point within the motion of a stride (typically defined at footstrike) to the walker's configuration at the same point in the next step, $x_{k+1}$. Let $\mathcal{S}$ define the Poincar\`e section, for which we have a Poincar\`e map $P: \mathcal{S} \rightarrow \mathcal{S}$ that maps one step to the next as $x_{k+1} = P(x_k)$.
The periodic orbit yields a fixed point $x^* = P(x^*)$ with $x^* \in \mathcal{O} \cap \mathcal{S}$, and stability of the orbit is equivalent to the stability of the Poincar\`e map which can be checked numerically \cite{morris2005restricted,wendel2010rank}.

\end{textbox}

The SLIP representation of walking has been primarily used for legged robots which have springs or series-elastic actuators (SEAs). Some of the earliest inclusions of compliant hardware on bipedal robots was with spring flamingo and spring turkey \cite{hunter1991comparative}. Later, the COMAN robot included passive compliance to reduce energy consumption during walking \cite{kormushev2011bipedal}, and the Valkyrie robot from NASA was the first full-scale humanoid robot to heavily use SEAs \cite{radford2015valkyrie}. 
Using inspiration from the SLIP morphology, Hurst designed the planar humanoid robot MABEL \cite{park2011identification} and the 3D bipedal robot ATRIAS \cite{grimes2012design, rezazadeh2015spring} to include series elastic actuation and thus return energy through impacts and shield the motors from impact forces at footstrike. One of the latest robots in this series, the Cassie biped (shown in \figref{fig:dynamic_model_contact_combined}(a)) also mechanically approximates SLIP design principles \cite{abateThesis}. Several running robots have specifically considered SLIP model principles in their mechanical design such as the ARL Monopod II \cite{ahmadi1999arl}, the CMU Bowleg Hopper \cite{brown1998bow}, and the Keneken hopper \cite{hyon2002development}.

\section{HYBRID SYSTEM MODELS OF BIPEDAL LOCOMOTION}
\label{sec:hybrid}

In the drive to obtain efficient legged locomotion and understand the stability thereof, 
researchers have adopted more dynamic paradigms for robotic locomotion which consider nontrivial impacts and periodic notions of stability.  
To formalize this perspective, it is necessary to consider hybrid system models of walking, which include both continuous (leg swing) and discrete (footstrike) dynamics.  This section discusses to key paradigms that leverage this framework: \textit{passive dynamic walking}, which exploits the natural hybrid dynamics of the system to obtain efficient walking, and \textit{hybrid zero dynamics} which uses actuation to achieve model reduction and thereby synthesize stable walking gaits. 

\subsection{Passive Dynamic Walking} \label{sec:passive_dynamic}


\begin{textbox}[b]
\section{Hybrid Dynamical Systems} \label{sidebar:hybrid_system}
A \textit{hybrid dynamical system}, used to model a walking robot \cite{ames2011human}, is defined as the tuple:
\begin{equation*}
  \label{eq:hybrid-dynamical-system}
  \HybridSystem = (\DirectedGraph,\Domain,\Guard,\ResetMap,F)
\end{equation*}
\begin{itemize}
  \item $\DirectedGraph = \{\Vertex,\Edge\}$ is a \emph{directed cycle} specific to the desired walking behavior, with $\Vertex$ the set of vertices and $\Edge$ the set of edges, $e = (v_{\rm s} \to v_{\rm t}) \in E$ with $v_{\rm s},v_{\rm t} \in V$, in the cycle.
  \item $\Domain=\{\Domain_{v} \}_{v \in V}$ is the set of \emph{domains of admissibility}. 
    Each domain $\Domain_\vi$ can be interpreted as the set of physically realistic states of the robot.
  \item $\Guard=\{\Guard_{\ei}\}_{\ei \in  E}$ is the set of \emph{guards}, with $\Guard_{\ei} \subset \Domain_{v_{\rm s}}$ which form the transition points from one domain, $\Domain_{v_{\rm s}}$ to the next in the cycle: $\Domain_{v_{\rm t}}$.
  \item $\ResetMap = \{\ResetMap_{\ei}\}_{\ei \in  E}$ 
      is the set of \emph{reset maps}, 
      $\ResetMap_{\ei} : \Guard_{\ei} \subset \Domain_{v_{\rm s}} \to \Domain_{v_{\rm t}}$ from one domain to the next. The reset map gives the post impact state of the robot: $x^+ = \ResetMap_{\ei}(x^-)$. 
  \item $F = \{f_v\}_{v \in V}$ is a set of \emph{dynamical systems} where $\dot{x} = f_v(x)$ for coordinates $x \in D_v$, i.e., of the form given in Equation \ref{eq:manip_nl_eom} with $u = 0$.
\end{itemize}

\end{textbox}



Some of the first work to study hybrid systems for the purposes of synthesizing walking were within the field of \textit{passive dynamic walking}. 
Tad McGeer \cite{mcgeer1990passive, mcgeer1990passiveknees} introduced several passive walking robots that could ambulate down small declines when started from a reasonable initial condition. While these early bipeds were completely passive and relied on gravity, several bipeds were built to demonstrate that simple actuators could substitute for gravitational power and compensate for disturbances. Small electric actuators were used for the Cornell walkers \cite{collins2001three,collins2005bipedal,bhounsule2009cornell} and the MIT learning biped \cite{tedrake2004stochastic,tedrake2004actuating}, while the Delft biped instead used a pneumatic actuator at the hip \cite{anderson2005powered,wisse2007delft}. 
Controlled symmetries \cite{spong2005controlled} and geometric reduction \cite{ames2007geometric} has been used to extend these ideas to actuated robots and 3D walking.
Actuated environments have also been used to excite walking on passive robots \cite{reher2020passive}. 
Because of the care taken in mechanical design, these robots could all operate without sophisticated real-time calculations---though at the cost of diminished control authority. 

The governing equations of motion for passive dynamic robots are nonlinear, and correspond to the continuous dynamics derived in \eqnref{eq:eom_continuous} rather than using an approximate (or reduced order) model. They are also \textit{hybrid}, meaning they consist of both continuous and discrete nonlinear dynamics.  A definition of the hybrid representation of the dynamics of walking is summarized in the sidebar ``Hybrid Dynamical Systems'', where the key element that determines the behavior is a directed cycle of
continuous domains.

\subsubsection{Discrete Dynamics: Impacts and Poincar\`e Maps}

\begin{figure*}[t!]
\centering
	\includegraphics[width=0.9\columnwidth]{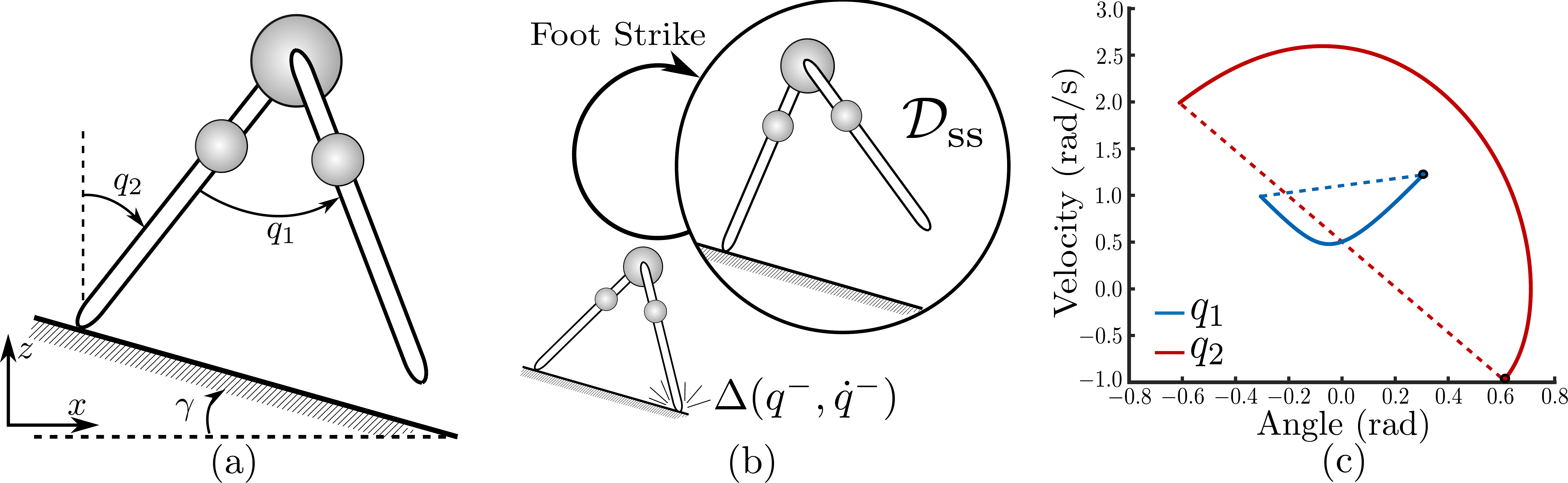}
    \caption{A canonical example of passive dynamic walking is the compass biped. (a) 
	The biped and its configuration on the slope. (b) Directed graph of the corresponding hybrid dynamical system.
	(c) 
	Shown is a closed limit cycle for the biped walking down a $5$ degree slope, implying stable walking.}
	\label{fig:passive_dynamic}
\end{figure*}

An inherent feature of dynamic walking is that the robot is moving quickly through the environment. 
Thus 
the resulting motions cannot be slow enough for the feet to approach the ground with negligible velocity; impacts with the ground, therefore, become an important consideration in dynamic walking. Formally accounting for 
impacts 
underlies the 
basis for 
hybrid dynamical 
locomotion models 
\cite{westervelt2018feedback, grizzle2014models, ames2011human}. 
Impacts during walking 
occur when the non-stance foot strikes the ground.  Concretely, consider the vertical distance (height) of a contact point (foot) above the ground: $H_e(x)$.  Impacts occur when the system reaches the \emph{switching surface} of the \emph{guard}: 
\begin{align}
    S_e = \{ x \in X ~ | ~ H_e(x) = 0, \dot{H}_e(x) < 0 \}.
\end{align}
where this surface is also a \emph{Poincar\`e section} that will be used to construct the \emph{Poincar\`e map.} At each transition, the new initial condition is determined through the \emph{reset map}:
\begin{align}
    \begin{bmatrix} q^+ \\ \dot{q}^+ \end{bmatrix} = 
    \begin{bmatrix} \mathcal{R} q^- \\ \mathcal{R} \Delta_{\dot{q}}(q^-) \dot{q}^- \end{bmatrix} = \Delta(q^-,\dot{q}^-), \label{eq:delta_impact}
\end{align}
where $\mathcal{R}$ is a relabeling matrix \cite{westervelt2018feedback,ames2012first}, which ``flips'' the stance and non-stance legs.  Here $\Delta_q$ describes the change in velocity that occurs at impact, and is typically calculated using the assumption of a perfectly plastic impact \cite{PFGL96, hurmuzlu1994rigid}.  Note that determining and utilizing more complex impact models is an open problem. 
In real life, impacts are not truly instantaneous and do not always achieve stiction \cite{or2014painleve}. Situations with multiple impacts can arise \cite{liu2008frictionless} leading to Zeno behaviors \cite{or2010stability,lamperski2012lyapunov,ames2011characterizing}, or slippage \cite{marton2007modeling, ma2019dynamic}.

A canonical example of passive dynamic walking is an unactuated compass biped walking down a slope of angle $\gamma$ \cite{goswami1998study}, shown in \figref{fig:passive_dynamic}(a). This robot consists of two kneeless legs each with a point-mass and a third mass at the hip. The directed cycle for the biped consists only of a single-support domain, with transition occurring at footstrike (shown in \figref{fig:passive_dynamic}(b)). The periodic nature of the stable walking behavior is best summarized by the phase portrait given in \figref{fig:passive_dynamic}(c), where there are discrete jumps occurring at impact. The stability of a cyclic gait is discussed in the sidebar ``Periodic Notions of Stability''. Once a fixed point $x^*$ has been found, we can examine a first order expansion of the Poincar\`e map:
\begin{align}
    P(x^* + \delta x) \approx x^* + \frac{\partial P}{\partial x}(x^*) \delta x,
    \qquad \mathrm{with} \qquad 
    P(x^*) = x^*, \qquad x^* \in \mathcal{O} \cap  S_e,
    \label{eq:poincare_linearized}
\end{align}
where the fixed point is exponentially stable if the magnitude of the eigenvalues of $\frac{\partial P}{\partial x}(x^*)$ are less than one \cite{perko2013differential,morris2005restricted,wendel2010rank}. 
This is straightforward to check numerically: one can construct a numerical approximation of successive rows by applying small perturbations to each corresponding state and then forward simulate one step to obtain $P(x^* + \delta x)$.

\begin{textbox}[b]
\section{Hybrid Control Systems} \label{sidebar:hybrid_control_system}
Rather than describing a passive hybrid system as in the sidebar ``Hybrid Dynamical Systems'', the incorporation of a feedback control allows for the realization of more advanced behaviors on complex actuated bipedal robots. We therefore define a 
\textit{hybrid control system} \cite{ames2011human,sinnet20092d} to be a tuple:
\begin{equation*}
    \HybridControlSystem = (\DirectedGraph,\Domain,\ControlInput,\Guard,\ResetMap,\emph{FG}).
\end{equation*}
\begin{itemize}
    \item $(\DirectedGraph, \Domain, \Guard, \ResetMap)$ are defined as in \eqnref{eq:hybrid-dynamical-system}.
    \item $\ControlInput=\{\ControlInput_{\vi}\}_{\vi \in V}$ is the set of \emph{admissible control inputs}.
    \item $\emph{FG} = (f_v,g_v)_{v \in V}$ the set of \emph{control systems}, $\dot{x} = f_v(x) + g_v(x) u$, as in  \eqnref{eq:manip_nl_eom}.
\end{itemize}

\end{textbox}

\subsection{Hybrid Zero Dynamics} \label{sec:hzd}

\begin{figure*}[t]
\centering
	\includegraphics[width=0.9\columnwidth]{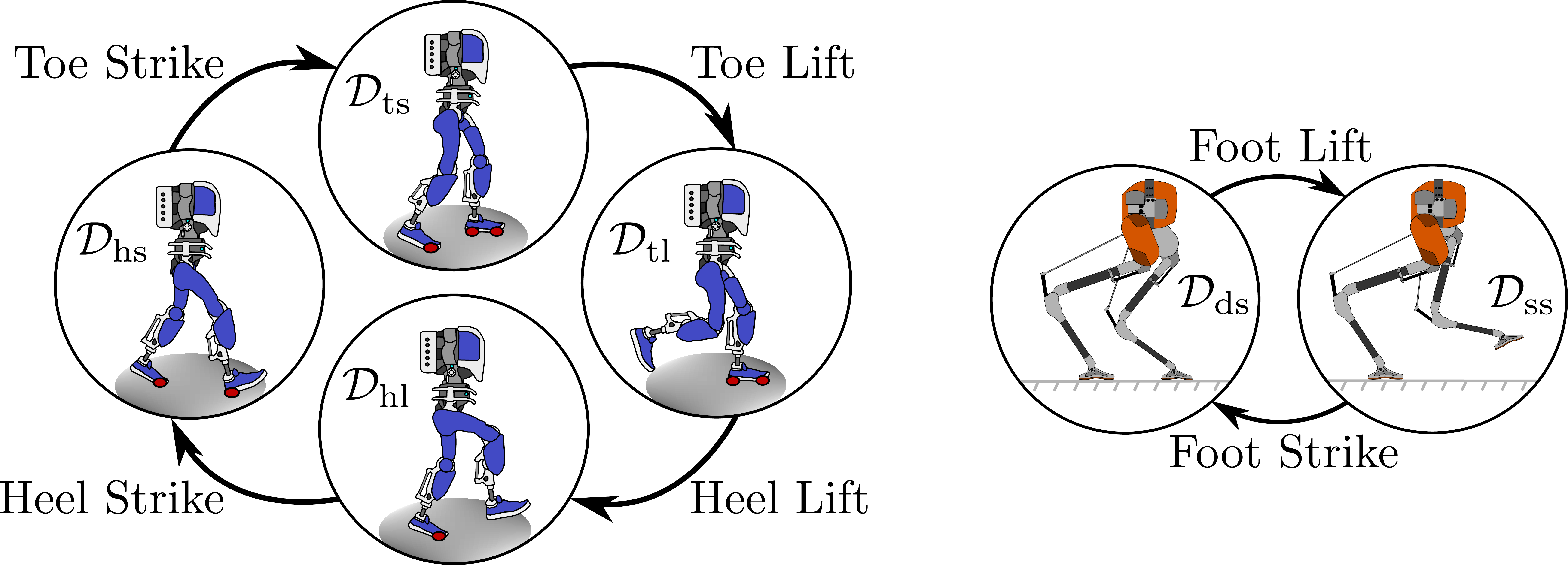}
	\caption{Examples of directed cycles for hybrid representations of two domain walking with compliance and four domain human-like robotic walking.}
	\label{fig:periodic_and_hybrid}
\end{figure*}

\begin{figure*}[t]
\centering
	\includegraphics[width=0.95\columnwidth]{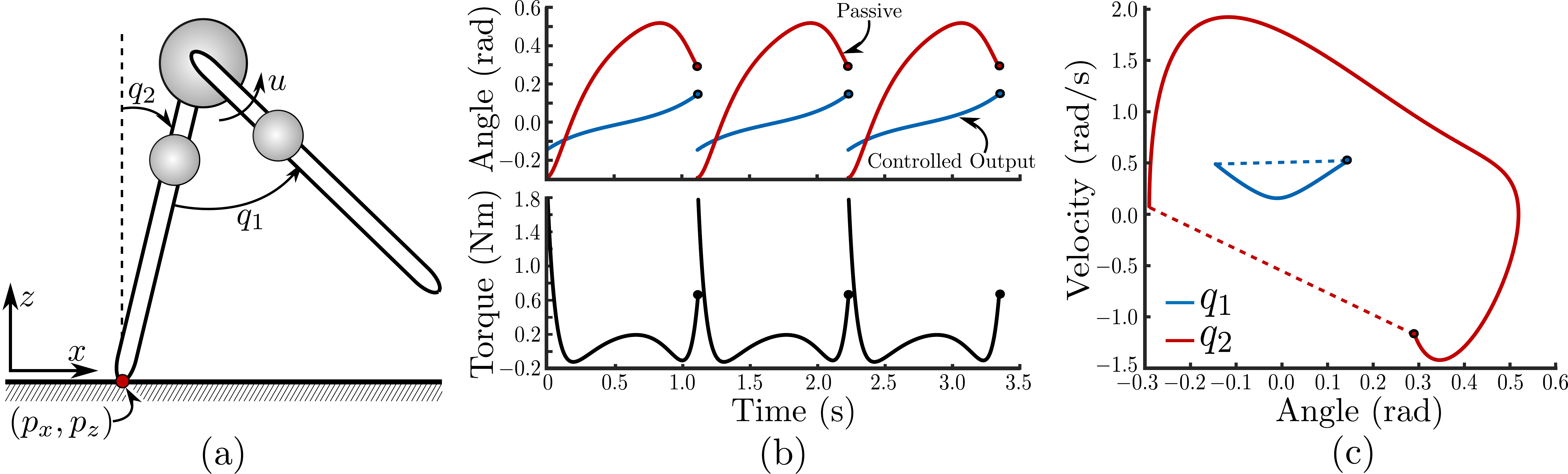}
	\caption{An example of HZD based control for a compass biped on flat ground. (a) The robotic configuration. (b) Joint trajectories and torques over three steps of stable walking. (c) The walking exhibits a stable limit cycle, with discrete jumps occurring at impact.}
	\label{fig:compass_hzd}
\end{figure*}

The method of \textit{hybrid zero dynamics} (HZD) leverages nonlinear feedback control design to induce stable locomotion on underactuated robots. Jessy Grizzle et al. \cite{westervelt2003hybrid,grizzle2001asymptotically,morris2009hybrid} introduced the concept and developed a set of tools which are grounded in nonlinear control theory to deal formally with the nonlinear and hybrid nature of dynamic walking (cf. the textbook \cite{westervelt2018feedback}). 
The basis of the HZD approach is the restriction of the full-order dynamics of the robot to a lower-dimensional attractive and invariant subset of its state space, the \emph{zero dynamics surface}, via \emph{outputs} that characterize this surface.
If these outputs are driven to zero, then the closed-loop dynamics of the robot is described by a lower-dimensional dynamical system that can be ``shaped''to obtain stability.
As was the case for uncontrolled hybrid models
generalizing hybrid dynamical systems, a hybrid control system (see sidebar) describes an actuated walking robot, leading to the notion of \emph{hybrid zero dynamics}.
The primary consideration which governs the overall locomotion problem is the specification of a directed cycle for the underlying hybrid (control) system. Because HZD incorporates feedback control, significantly more complex motions are possible.
Examples of directed cycles for 
dynamic walking behaviors 
are given in \figref{fig:periodic_and_hybrid}, 
illustrating
how 
domain specification is governed largely by the evolution of the 
contacts 
through the course of a step. The controlled compass walker \cite{goswami1998study} is presented in \figref{fig:compass_hzd} to provide a comparison to the passive dynamic walking. In this example it can be seen that torques applied at the hip are used to control the motion, while the robot walks with a stable limit cycle on flat ground.

\subsubsection{Virtual Constraints and Stabilization}  \label{sec:virtualconstraint}
Dynamic walking which leverages the fullbody dynamics must necessarily include specifications on how the robot should coordinate its limbs in a holistic fashion. 
To this end, and analogous to holonomic constraints, \textit{virtual constraints} are defined as a set of functions that regulate the motion of the robot to achieve a desired behavior \cite{westervelt2018feedback}. The term ``virtual'' comes from the fact that these constraints are enforced through feedback controllers instead of through physical constraints. 
Let $y^a(q)$ be functions of the generalized coordinates that are to be controlled, i.e., encoding the ``actual'' behavior of the robot and $y^d(t,\alpha)$ be the ``desired'' behavior where $\alpha$ is a matrix of real coefficients that parameterize this behavior. 
A B\'ezier polynomial is the most typical choice of representation for the desired outputs \cite{westervelt2018feedback} for computational reasons, though it has been shown that humans appear to follow spring-mass-damper type behavior \cite{ames2012first}.
Given actual $y^a$ and desired $y^d$ outputs, a \emph{virtual constraint} is their difference: 
\begin{eqnarray}
    \label{eq:outputs}
    y(q) := y^a(q) - y^d(\tau(q), \alpha),
\end{eqnarray}
with $\tau(q):\mathcal{Q}\rightarrow\mathbb{R}$ a \emph{parameterization of time} that is strictly increasing along periodic motions. 
Driving $y \to 0$ results in convergence of the actual outputs to the desired. 

To synthesize controllers, note that differentiating $y(q)$ along solutions to the control system in \eqnref{eq:manip_nl_eom} yields the \emph{Lie derivatives}: 
\begin{eqnarray}
\dot{y}(q,\dot{q}) = L_f y(q,\dot{q}), \qquad \ddot{y}(q,\dot{q}) = L_f^2 y_2(q,\dot{q}) + L_g L_f y(q,\dot{q}) u,
\end{eqnarray}
wherein $y(q)$ has vector relative degree 2 \cite{sastry2013nonlinear} if the matrix $L_g L_f y(q,\dot{q})$ is invertible.
From this, one obtains the following feedback linearizing controller: 
\begin{align}
    u^*(x) =  \left[L_g L_f y(x)\right]^{-1}\Big(-L_f^2 y(x) + \mu \Big) 
    \qquad \Rightarrow \qquad 
    \ddot{y}(q,\dot{q}) = \mu,
    \label{eq:FBL}
\end{align}
where $\mu$ is the auxiliary feedback component of the controller that can be chosen to stabilize the system.  In particular, one common choice is
$\mu = \frac{1}{\epsilon^2}K_{\mathrm{p}} y(x) + \frac{1}{\epsilon} K_\mathrm{d} L_f y(x)$, 
with $K_\mathrm{p}$, $K_\mathrm{d}$ feedback gains chosen so that the linear dynamics are stable, and $\epsilon>0$ a parameter used to amplify convergence to the desired motion; rendering the output dynamics exponentially stable.  Applying \eqnref{eq:FBL} with 
$\mu$ 
results in the closed-loop system: 
\begin{align}
\label{eq:eom_nonlinearcl}
    \dot{x} = f_{\mathrm{cl}}(x) = f(x) + g(x) u^*(x).
\end{align}
wherein for this system $y \to 0$ exponentially fast.

\begin{figure*}[b]
\centering
	\includegraphics[width=0.9\columnwidth]{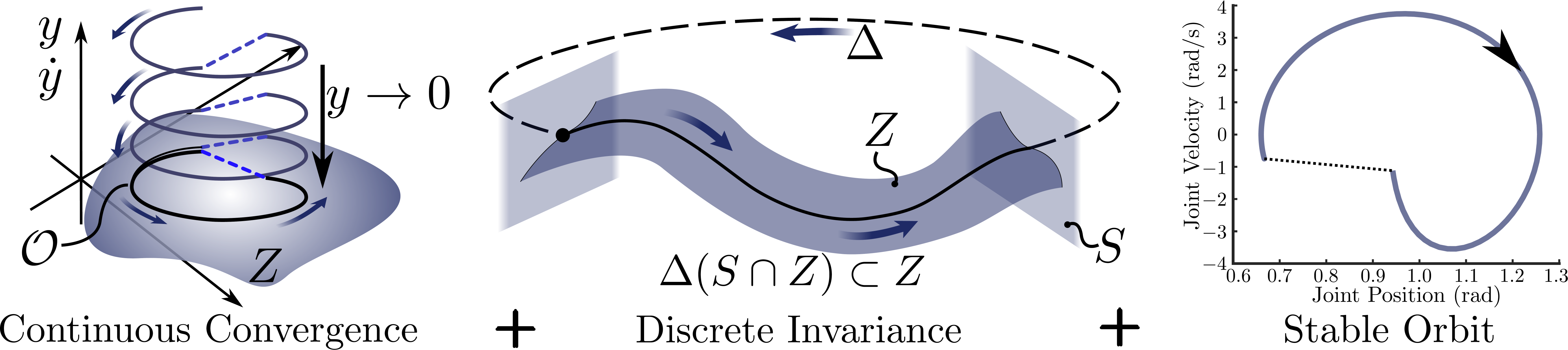}
	\caption{
	Key concepts 
	related to 
	hybrid zero dynamics: continuous convergence to a 
	zero dynamics surface $Z$, coupled with a hybrid invariance condition: $\Delta(Z \cap S) \subset Z$ to obtain stable periodic walking.}
	\label{fig:hzd_concept}
\end{figure*}

\subsubsection{Hybrid Invariance}  The feedback control law of \eqnref{eq:FBL}
can be synthesized for virtual constraints $y_v(q) = y^a_v(q) - y^d_v(\tau(q,t),\alpha_v)$ associated with each
domain $\Domain_v$, $v \in V$, and control system: $\dot{x} = f_v(x) + g_v(x)u$.  This renders the \textit{zero dynamics manifold} \cite{isidori1997nonlinear}:
\begin{align}
\label{eq:zerodyn}
    \mathbb{Z}_v = \{ (q,\dot{q}) \in \mathcal{D}_v ~  | ~ y_v(q) = 0 ,~   \dot{y}_v(q,\dot{q})  = 0 \},
\end{align}
forward invariant and attractive. Thus, the continuous dynamics in \eqnref{eq:eom_nonlinearcl} will evolve on $\mathbb{Z}_v$ given an initial condition in this surface. 
However, because the surface in \eqnref{eq:zerodyn} has been designed without taking into account the hybrid transition maps of \eqnref{eq:delta_impact}, the resulting walking cycle may not be invariant to impact. To enforce impact invariance, the desired outputs can be shaped through the parameters $\alpha_v$ in $y^d_v$ such that the walking satisfies the \textit{hybrid zero dynamics} (HZD) condition:
\begin{align}\label{eq:resetmapv}
 \Delta_e ( \mathbb{Z}_{v_{\rm s}} \cap S_{v_{\rm s}} ) \subset \mathbb{Z}_{v_{\rm t}}, 
 \qquad \forall ~ e = (v_{\rm s},v_{\rm t}) \in E,
\end{align}
imposed as a constraint on the states through impact (\eqnref{eq:delta_impact}). 

The overarching goal of these constructions is to provide a framework for the synthesis of dynamic walking gaits. 
In this context, for simplicity (and without loss of generality) assume a single domain $V = \{v\}$ wherein we will drop the subscript ``$v$.'' 
For the full-order dynamics, let $\phi_t^{f_{\mathrm{cl}}}(x_0)$ be the (unique) solution at time $t\geq 0$ with initial condition $x_0$. For a point $x^*\in S$ we say that $\phi_t^{f_{\mathrm{cl}}}$ is hybrid periodic if there exists a $T>0$ such that $\phi_T^{f_{\mathrm{cl}}}(\Delta(x^*))=x^*$. Further, the stability of the resulting hybrid periodic orbit, $\mathcal{O} = \{ \phi_t^{f_{\mathrm{cl}}}(\Delta(x^*)):0 \leq t \leq T \}$, 
can be found by analyzing the stability of the Poincar\'{e} map, wherein $x^*$ is a fixed point, as previously presented in \eqnref{eq:poincare_linearized}. The main idea behind the HZD framework is, due to the hybrid invariance of $\mathbf{Z}$, if there exits a stable hybrid periodic orbit, $\mathcal{O}_{\mathbf{Z}}$, for the reduced order \emph{zero dynamics} evolving on $\mathbf{Z}$, i.e., the restriction of $f_{\rm cl}$ to $\mathbf{Z}$, then $\mathcal{O}_{\mathbf{Z}}$ is a stable hybrid periodic orbit for the full order dynamics in \eqnref{eq:zerodyn} \cite{westervelt2018feedback}. 
A visualization of the 
components of HZD walking design is given in  \figref{fig:hzd_concept}.

In the case of robots which have feet, as is the case for many humanoid robots, one can extend the concept of HZD to modulate the forward velocity of the robot \cite{ames2014human}.  In particular, one can generalize HZD through 
a velocity modulating output: $y_1(q,\dot{q}) = y_1^a(q,\dot{q}) - v^d$, with $v^d$ the desired forward velocity.  
One can augment the original virtual constraints $y(q)$ with this new (relative $1$ degree) output.  
The partial hybrid zero dynamics surface $\mathbb{PZ}$ is, again, given as in \eqnref{eq:zerodyn} where the term ``partial'' is used since this surface does not require the output $y_1$ to be zero.  Partial hybrid zero dynamics (PHZD) is the condition: $\Delta_e ( \mathbb{PZ} \cap S ) \subset \mathbb{PZ}$.  In the case of full actuation the existence of hybrid periodic orbit, $\mathcal{O}_{\mathbf{PZ}}$ in $\mathbf{PZ}$, is guaranteed implying the existence of a hybrid periodic orbit for the full order dynamics.  Thus, PHZD implies the existence of a stable gait for fully actuated robots.

\begin{figure*}[b]
\centering
	\includegraphics[width=1\columnwidth]{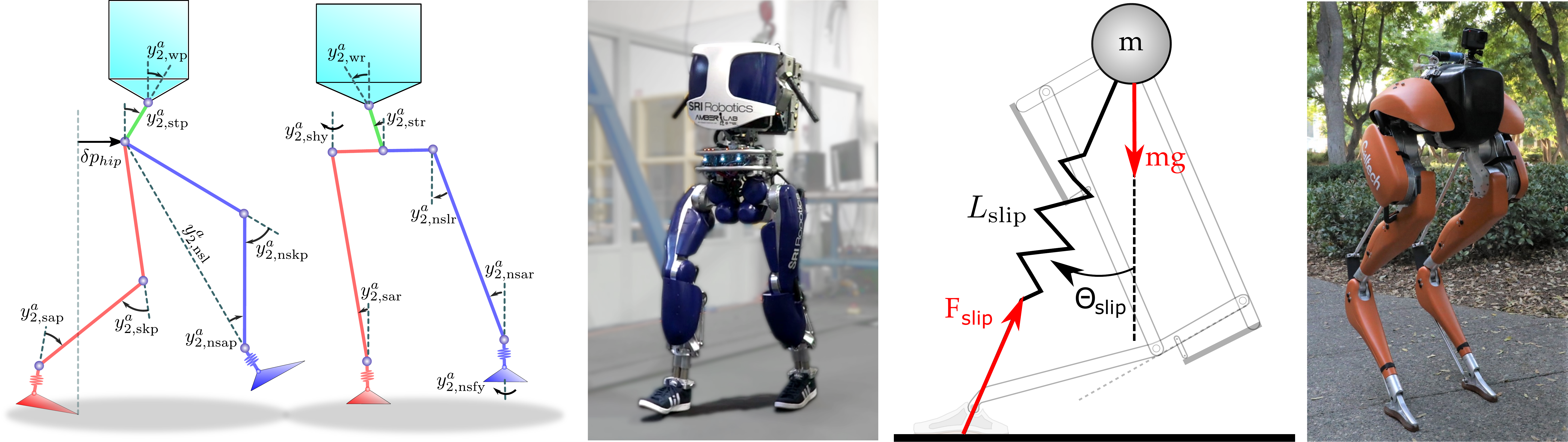}
	\caption{
	A visualization of the human-like outputs \cite{ames2014human} applied to 
	DURUS 
	(left), 
	and a depiction of how the physical morphology of 
	Cassie 
	follows principles from SLIP models (right). The robot then has passive dynamics which can be embedded within the HZD framework via output selection. 
	}
	\label{fig:hzd_output_selection}
\end{figure*}

\subsubsection{Application of HZD}  
In the context of robotic implementations, HZD has proven successful in realizing a wide variety of dynamic behaviors. Many of the early uses of the method were on point-footed robots which were restricted to the sagittal plane. The first robot used to study HZD was the Rabbit biped \cite{chevallereau2003rabbit}, followed later by MABEL \cite{grizzle2009mabel} and AMBER $1$ \cite{yadukumar2013formal}. The ability of (P)HZD to handle multidomain behaviors led to its use on more complex planar bipedal robots such as ATRIAS \cite{hamed2013event,ramezani2014performance}, AMBER $2$ \cite{zhao2014human} and AMBER $3$M \cite{ambrose2017toward}. 
New challenges appeared while extending the method of HZD from planarized robots to 3D robots, which exhibit additional degrees of underactuation. Control of fully actuated humanoids was demonstrated on a small-scale example with a NAO robot \cite{ames2012dynamically} via PHZD, while point-footed 3D walking with HZD was first shown at the University of Michigan with the MARLO biped \cite{buss2014preliminary}. 
At the DARPA Robotics Challenge, 
the humanoid DURUS (shown in \figref{fig:hzd_output_selection}(a)) was featured in an efficiency walk-off \cite{reher2016durusrealizing} where it demonstrated the first sustained humanoid HZD walking---over five hours continuously. DURUS went on to exhibit the most efficient walking on a humanoid to date, while performing human-like multicontact behaviors and managing significant underactuation \cite{reher2016durusmulticontact}. 
The method has been extended to powered prosthetic walking \cite{zhao2016multicontact,zhao2017first,zhao2017preliminary} and to exoskeletons which can walk for patients with paraplegia \cite{harib2018feedback,gurriet2018towards}. 
The use of springs in locomotion has also proven useful in the development of dynamic walking behaviors, though it presents additional challenges both mathematically and in practice. The notion of compliant hybrid zero dynamics was introduced in the late $2000$'s \cite{poulakakis2009stabilizing}, and was later expanded upon to obtain compliant robotic running \cite{sreenath2013embedding}. One of the latest robots to successfully demonstrate stable HZD walking is the Cassie biped, shown in \figref{fig:dynamic_model_contact_combined}(a), where it can be seen that the robot exhibits underactuated feet and passive springs in the legs. Dynamic walking on Cassie has been successfully realized on hardware both by planning under the assumption of sufficient rigidity in the legs to ignore compliant elements \cite{gong2019feedback}, and for walking which considers the passive compliance in the zero dynamics \cite{reher2019dynamic}.

%
\begin{figure*}[b]
	\centering
	\includegraphics[width= 1.0 \columnwidth]{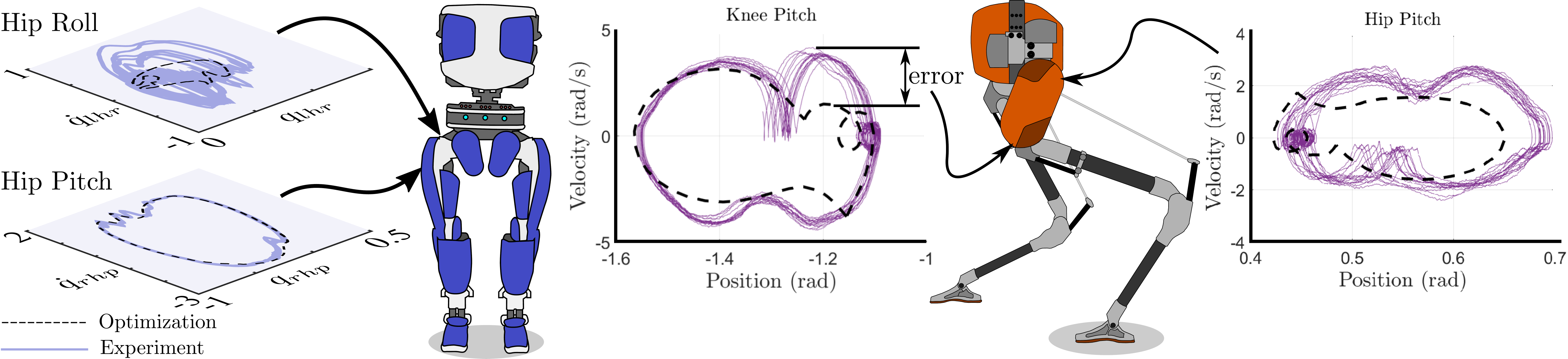}
	\caption{ 
	An example of 
	HZD periodic orbits for dynamic walking on hardware with 
	the 
	DURUS (left) and Cassie (right)
	robots. 
	The nominal periodic walking motions resulting from the optimization (\eqnref{eq:opteqs}) are shown as a dashed line, superimposed on 
	traces of 
	experimental data. }
	\label{fig:pd_experiment}
\end{figure*}

\begin{summary}[Experimental Highlight: Hybrid Zero Dynamics]
%
%
We highlight the application of HZD by considering its experimental realization on hardware. 
Leveraging PHZD on DURUS and HZD on Cassie, the result is stable periodic orbits---both in simulation and experimentally---as illustrated in \figref{fig:periodic_and_hybrid}. 
The evolution of the dynamic walking motion 
is tied to the morphology of the robot, with the humanoid DURUS exhibiting human-like heel-toe walking 
\cite{ames2014human,reher2016durusmulticontact} and the Cassie biped leveraging a domain structure and outputs which correspond to the 
SLIP inspired mechanical design \cite{reher2019dynamic}. 
In these specific examples, the virtual constraints chosen are shown in \figref{fig:hzd_output_selection}. 
This demonstrates one of the benefits of HZD: the ability to chose virtual constraints to formally encoded reduced order models for complex robots and correspondingly ``shape'' the zero dynamics surface to render it stable. 
\end{summary}

\section{MOTION GENERATION FOR DYNAMIC BIPEDAL LOCOMOTION} \label{sec:motion_planning}
Throughout the previous section, we outlined how the locomotion problem is fundamentally different than traditional approaches to modeling fixed-base robots. It is because of this inherent complexity that virtually all approaches to realize dynamic walking must transcribe the locomotion problem into a motion planner which can handle the various constraints naturally imposed on the problem. While several of the more classical walking paradigms offer simple solutions to conservative walking, there has been a push over the last two decades towards leveraging optimization to obtain increasingly dynamic maneuvers.

\subsection{Step Planning with Linear and Reduced-Order Models} \label{sec:step_plan_linear}
Often for the simplest models of walking, such as traditional ZMP and 
LIPM versions of CP, 
the linear dynamics of the restricted system yield straightforward approaches to planning the motion of the COM. The walking characterized by these linear models often implicitly satisfy quasi-static stability assumptions, ultimately allowing a control designer to 
decouple the high-level step planner and low-level balance 
controllers \cite{2016:Humanoids-Stumpf}. In this vein, Kajita \cite{kajita2003biped} introduced the jerk of the COM 
as an input 
controlled by a discrete LQR controller with preview action \cite{nishiwaki2009online} to plan ZMP trajectories for predefined footsteps. However, predefining the motions of the ZMP or footholds is not always necessary or desirable. 

%
If planners for these simple models could instead be performed online, then the robot may be able to mitigate issues related to reactivity. 
Weiber \cite{wieber2006trajectory} proposed using linear trajectory-free \textit{model predictive control} (MPC) as a method for explicitly handling the constraints imposed by the ZMP approach of \secref{sec:zmp} while continuously re-evaluating the walking path. Stephens \cite{stephens2010push} presented the use of MPC for push recovery and stepping on the SARCOS humanoid, which could be extended to obtain walking behaviors. The example shown in \figref{fig:zmp_lip_example}(c) visualizes the result of this MPC approach applied to LIPM robotic walking. It has also been shown how optimization and model predictive control can extend the notions of capture point to viable regions on which the biped can step \cite{wieber2008viability}, or how push recovery can be planned over a horizon of multiple steps \cite{koolen2012capturability}.
Despite the ability of these planners to adapt online, they cannot handle the discrete dynamics associated with footstrike, and demand near-zero impact forces \cite{huang1999high}. This rules out the nontrivial impacts which are naturally associated with dynamic walking. It also difficult to provide a priori guarantees on whether any given reduced-order plan is feasible to execute on the full-order dynamics. Such methods typically use inverse kinematics \cite{KaMaHaKaKaFuHi03}, or inverse dynamics \cite{nagasaka1999dynamic} sometimes in an operational-space formulation \cite{khatib1987unified} to compute the full-order control inputs at each instant. Solving such near-term inverse problems does not imply that future inverse problems in the trajectory will be feasible, which requires additional planning \cite{hebert2015mobile,zucker2015general}.

\subsection{Nonlinear Optimization for Gait Generation} \label{sec:nonlinear_planning}

As a result of the 
rapid developments within the trajectory optimization community, researchers began to move towards utilizing nonlinear dynamic gait optimizations rather than relying on the constraints imposed by linear modeling assumptions. 
The use of nonlinear optimization, i.e., numerical approaches, to generate stable walking behaviors on bipeds not a new concept \cite{chow1971studies,channon1992derivation}, though computational limitations were a considerable hindrance towards generating motions on 3D robots. Computation power finally increased sufficiently throughout the mid $2000$'s to begin handling 3D dynamic walking behaviors \cite{mombaur2009using}. 

\subsubsection{Open-Loop Optimization}
The simplest application of nonlinear optimization to walking can be found in \secref{sec:passive_dynamic}, wherein passive dynamic walking relies on the generation of fixed points associated with periodic orbits of a hybrid dynamical system. 
This naturally lends itself to numerical approaches for the optimization of \textit{open-loop} stable periodic motions \cite{mombaur2005open}, since passive dynamic walkers do not have any actuators to consider. 
The use of open-loop optimization to generate feasible motions for actuated robots are a natural extension of approaches used throughout the field of trajectory optimization, where the planning problem is seen as ``decoupled'' from the feedback control applied to the actual robot \cite{dai2014whole} and ``approximately optimal'' solutions are often sufficient. 
Further, in recent years, the application of advanced trajectory optimization methods such as \textit{direct collocation} have allowed the optimization of the full body dynamics of \eqnref{eq:eom_continuous} to be more computationally tractable, sparking a growing interest in considering the fullbody dynamics of robot in the planning problem. 
%
%
For instance, in order to control the open-loop trajectory that results from the direct collocation optimization, a classical linear quadratic regulator (LQR) based feedback controller can be constructed to stabilize the resulting trajectory obtained for the constrained dynamical system \cite{posa2016optimization}. In this type of approach, the walking problem can be viewed as generating sequences of footholds for the nonlinear centroidal dynamics given in \eqnref{eq:centroid_dynamics} 
\cite{herzog2015trajectory,kuindersma2016optimization} or with respect to the full Lagrangian system given in \eqnref{eq:eom_continuous} \cite{posa2014direct}.  Complementary Lagrangian systems \cite{moreau1966quadratic} formed the basis of the approach in \cite{posa2014direct}, which allowed the optimizer to find walking behaviors without a priori enumeration of the type and order of contact events. Open-loop trajectory optimization has also been used to satisfy ZMP conditions in a nonlinear fashion \cite{denk2001synthesis}, which considerably improved the dynamical nature of the conservative walking presented in \secref{sec:zmp}. 

\subsubsection{Closed-Loop Optimization}
\begin{figure*}[t]
	\centering
	\includegraphics[width= 0.95 \columnwidth]{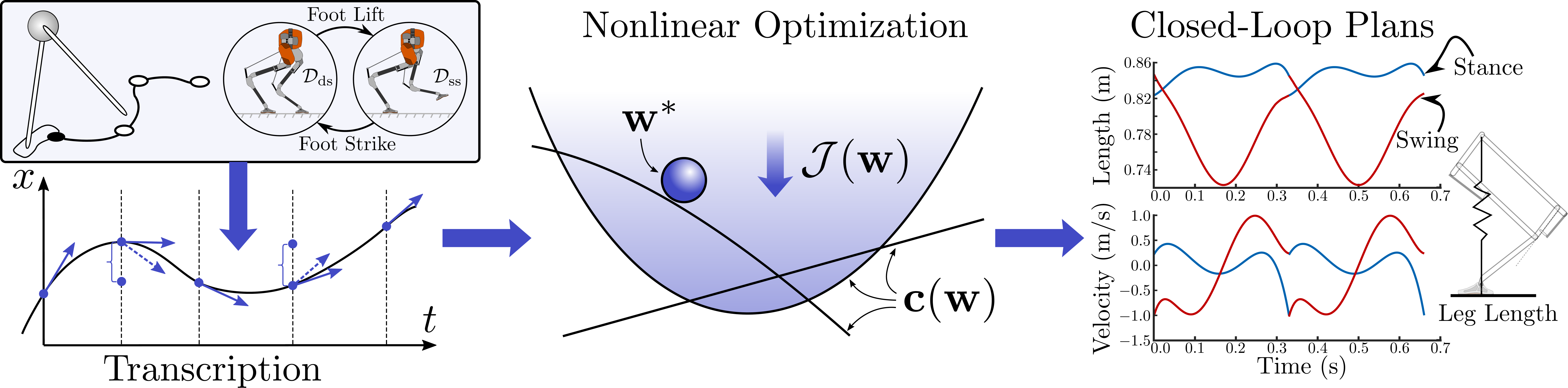}
	\caption{ A conceptual illustration of how locomotion models must be first transcribed into appropriate representations for use with nonlinear programming approaches in order to yield dynamically stable closed-loop plans for bipedal robots. }
	\label{fig:optimization_illustration}
\end{figure*}

While the preceding nonlinear optimization approaches do consider the fullbody dynamics of the robot, it is not always desirable to apply feedback controllers to stabilize an approximately optimal open-loop plan.  
Rather, it is often beneficial to couple the gait generation and controller synthesis problems into a single framework: \textit{closed-loop optimization}. 
This allows, among other things, for the generation of \textit{provably stable} walking behaviors which simultaneously satisfy the constraints on the system from admissible configurations to torque bounds. 
This idea forms the basis of designing walking gaits with the HZD method introduced in \secref{sec:hzd}, where feedback control is used to generate provable stable periodic orbits. A visual summary of this section is given in \figref{fig:optimization_illustration}. 
By applying these closed-loop feedback strategies in the optimization problem, ambiguous contact sequences are no longer possible \cite{hereid20163d} and must be prescribed according to the directed cycle which governs the underlying hybrid system (see \figref{fig:periodic_and_hybrid}).  
Doing so allows for one to enforce physical feasibility constraints, e.g., unilateral contact, in conjunction with the synthesis of controllers that guarantee stability. 

In the context of HZD methods, with the formal constructions of the zero dynamics (\eqnref{eq:zerodyn}) and hybrid invariance (\eqnref{eq:resetmapv}) defined, the problem of finding stable dynamic walking can be transcribed to a nonlinear programming (NLP) problem of finding a fixed point $x^*$ and set of parameters $\alpha = \{\alpha_v\}_{v \in V}$ parameterizing the virtual constraints of \eqnref{eq:outputs}. 
The optimization problem is performed over one step cycle, e.g., footstrike to footstrike, with a constraint imposed such that when the discrete impact (\eqnref{eq:delta_impact}) is applied to the terminal state so that it satisfies the hybrid invariance condition of \eqnref{eq:resetmapv}. It is also critical that the motions respect the limitations of the physical system such as the friction cone (\eqnref{eq:cone_friction}), actuator limits, and joint limits. 
These constraints can be directly placed into a NLP problem that can be solved by a standard optimization solver: 
\begin{align} 
    \label{eq:opteqs} 
	\mathbf{w}(\alpha)^* = \underset{\mathbf{w}(\alpha)}{\mathrm{argmin}} &\hspace{3mm} \mathcal{J}(\mathbf{w}(\alpha)) \\
	\mathrm{s.t.} 		&\hspace{3mm}  \text{Closed\ loop\ dynamics: \eqnref{eq:eom_nonlinearcl}}  \notag \\
				  		&\hspace{3mm}  \text{HZD\ condition: \eqnref{eq:resetmapv}}  \notag\\
				  		&\hspace{3mm}  \text{Physical feasibility (e.g. \eqnref{eq:cone_friction})}   \notag
\end{align}
where $\mathbf{w}(\alpha)\in\mathbb{R}^{N_w}$, with $N_w$ being the total number of optimization variables and here we made the dependence on the parameters, $\alpha$, that dictate the closed loop dynamics explicit. 
With the goal of achieving dynamic and efficient walking, a common objective is to minimize the \textit{mechanical cost of transport} (M-COT) of the walking gait through the cost \cite{reher2016durusrealizing,hereid20163d}.
In classical HZD implementations, the candidate solutions were found via 
single shooting formulations \cite{westervelt2003hybrid,ames2014human}, where the decision variables are the fixed point states $x^*$ and the 
output coefficients $\alpha$. Because single shooting optimizations are notoriously sensitive to poor initial conditions, multiple shooting 
was
also explored \cite{hereid2015hybrid}, with the eventual development of direct collocation formulations \cite{hereid2018dynamic} which would become the most successful to date. The FROST optimization package \cite{hereid2017frost} was developed based on these successes as an open-source package to transcribe HZD locomotion into a direct collocation problem. 
While HZD optimization problem determines one stable walking orbit, it has been shown that one can expand the range of motions a robot can perform through systematic optimization to build libraries of walking parameters \cite{da20162d}. Reinforcement learning has also been used to handle robust transitions for different speeds or unknown terrain height disturbances \cite{da2017supervised}.


\begin{summary}[Experimental Highlight: Closed-Loop Optimization]
%
%
The use of closed-loop optimization for HZD behaviors yields a set of outputs which coordinate the motion of the robot. This is shown as the output of the process in \figref{fig:optimization_illustration}, where the nonlinear optimization problem has provided outputs which yield orbital stability for compliant HZD \cite{reher2019dynamic}. The outputs which are shown directly correspond to the SLIP-like morphology of the robot, emphasized in \figref{fig:hzd_output_selection}. 
Reference trajectories can be shaped by the cost to yield desirable gait characteristics, such as efficiency on DURUS, 
or for minimizing torque and extraneous movement on Cassie to obtain behaviors which leverage the compliance for propulsion.
\end{summary}
\section{FEEDBACK CONTROL AND MOTION REGULATION} \label{sec:regulation}

While the dynamic walking paradigms introduced throughout the previous sections generate stable walking motions in simulation, their actual implementation requires the deployment of real-time feedback controllers capable of achieving the desired motions. 
As described in \secref{sec:fullbody_model}, dynamic walking robots involve a high level of complexity in the form of nonlinearities and tightly coupled equations of motion which must be considered. 
%
%
In the case when locomotion has been planned using a simplified model (\secref{sec:fullbody_model}, the spatial geometry of the robot must be translated into joint angles which can be controlled. 
Even with a full-order hybrid model (\secref{sec:hybrid}) and closed-loop optimization (\secref{sec:motion_planning}), 
controllers must be synthesized in order to track these desired motions in practice.  
This section describes feedback controllers and motion regulators that allow for the translation of dynamic walking in simulation to be realized on real-world hardware platforms. 

\subsection{Controllers for Tracking Designed Motions}


The simplest control scheme for determining motor torques is Proportional-Derivative (PD) control \cite{ziegler1942optimum}.  The strongest argument for using this approach is the sheer simplicity in its implementation and the intuitive physical meaning with respect to tuning.  Consider desired positions and velocities, $q^d$ and $\dot{q}^d$ (and possibly functions of time), either obtained from inverse kinematics for reduced order walking models or the output of a optimization problem.  A feedback controller can be applied at the joint level:
\begin{align} \label{eq:PD_joint}
    u = - K_\mathrm{p} (q^\mathrm{a} - q^\mathrm{d}) - K_\mathrm{d} (\dot{q}^\mathrm{a} - \dot{q}^\mathrm{d}),
\end{align}
generating desired torques (or currents) that are tracked at the motor controller level at a fast loop rate.
in the case of underactuated robots and/or virtual constraints (see \secref{sec:virtualconstraint}), one can consider outputs of the form:
$$
y(q) = y^a(q) - y^d(\tau(q),\alpha) \qquad \mathrm{or} \qquad 
y(q,t) = y^a(q) - y^d(\tau(t),\alpha),
$$
where the time-based variant is often considered in practice, especially in the case of 3D walking and running, due to imperfect sensing of $\tau(q)$ wherein it is replaced by the more robust signal $\tau(t)$ \cite{ma2017bipedal,kolathaya2016time}. 
Let $q_m$ represent the joints with actuators, then the PD controller can be applied in the Cartesian (or output) space:
\begin{align} \label{eq:PD_output}
    u = - Y(q)^{-1} (K_\mathrm{p} y + K_\mathrm{d} \dot{y})  \qquad \text{  or  } \qquad  u = - Y(q)^{T} (K_\mathrm{p} y + K_\mathrm{d} \dot{y}),
\end{align}
where $Y(q) := \frac{\partial y^a } { \partial q_m} (q) $ is the Jacobian of the Cartesian task or output with respect to the actuated joints, and $K_\mathrm{p}, K_\mathrm{d}$ are the PD gain matrices. 
This style of feedback control has been used to enforce the behaviors of every locomotion paradigm detailed in \secref{sec:models} and \secref{sec:hybrid} at some point in time. 

For underactuated dynamic walkers whose motions have been planned with virtual constraints, simply tracking the outputs with a well tuned PD controller is sometimes sufficient to achieve walking on hardware \cite{westervelt2004experimental,grizzle2009mabel,da20162d,reher2016durusmulticontact,reher2019dynamic}, and even running \cite{ma2017bipedal}.
This is because the trajectories (or outputs) \textit{implicitly} encode the dynamic behavior and stability constraints, even if it requires different torques on the actual robot to achieve these behaviors. In addition, because dynamic behaviors are often rendered stable through this behavioral encoding while satisfying appropriate physical constraints, 
almost all passive dynamic and HZD walkers to date have opted not to include load cells in the feet -- as feedback control of these quantities is not necessary for stability.
An example of PD controllers applied in experiment to two 3D bipedal robots is given in \figref{fig:pd_experiment}, where it can be seen that although the motions do not track the designed motions perfectly they do form a closed orbit---implying stable walking. 

In the context of reduced order models, plans for ZMP and CP have typically considered a point-mass representation of the robot under which whole-body momentum and force regulation becomes an important concern when developing feedback controllers for implementation. This has led to a variety of approaches which concurrently regulate the COM movement via some PD feedback element in combination with control of the whole-body momentum \cite{kajita2003resolved,popovic2004zero,lee2010ground,koolen2016design} and tracking of desired force interactions \cite{fujimoto1996proposal,saab2013dynamic}. 

When the dynamics of the system are well known, it is often beneficial to leverage them in the feedback control design. One of the classical methods which was used for exploring this in the context of bipedal robots is \textit{computed-torque control}, which considers an inner nonlinear compensation loop, and the design of an auxiliary control feedback \cite{tzafestas1996robust,park1998biped}:
\begin{align}
    u = D(q)\left(\ddot{q}^* - K_p (q - q^{\mathrm{d}}) - K_d(\dot{q} - \dot{q}^\mathrm{d})\right) + H(q,\dot{q}),
\end{align}
where $\ddot{q}^*$ is the nominal system acceleration. 
Note that this is mathematically equivalent to feedback linearization, as given in Equation \eqref{eq:FBL} (see \cite{ames2013towards}). 
Although both the standard PD controller and computed-torque approach can overcome minor disturbances, they are often not sufficient to formally ensure the stability or yield the performance that dynamic walking requires. 
This motivates the use of a controller which can provide good tracking performance while leveraging the robotic model. The remainder of this section will explore several of the approaches which have been successful in the feedback control of bipedal robots, and how these can be extended to provide formal stability guarantees. 

\begin{figure*}[b]
	\centering
	\includegraphics[width= 1.0 \columnwidth]{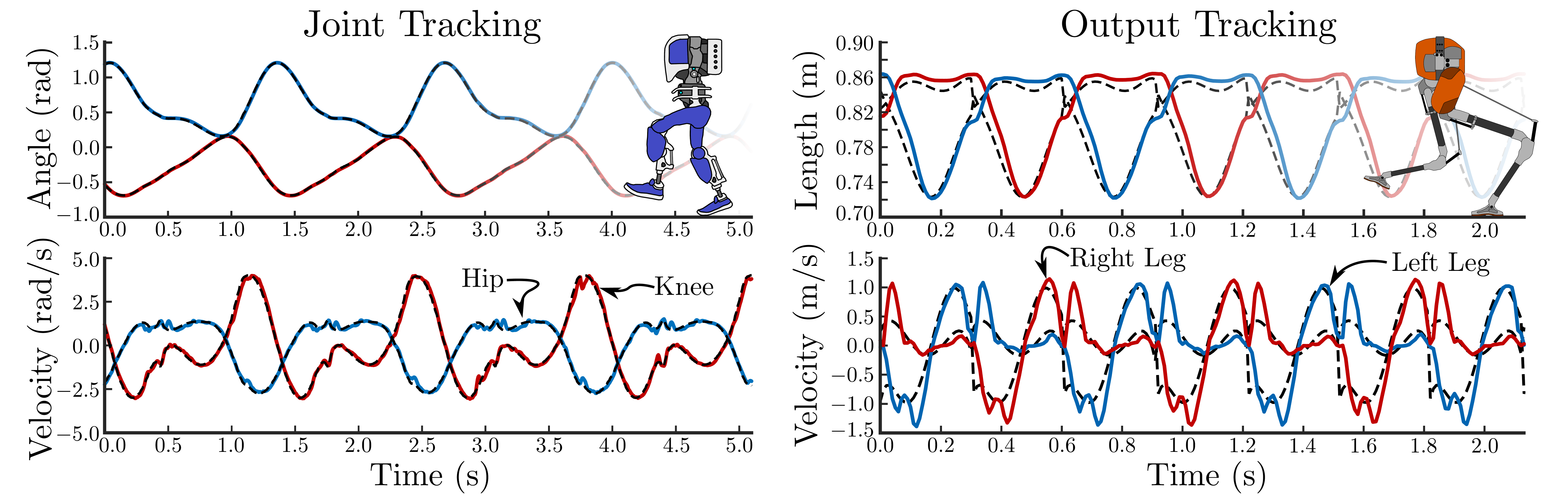}
	\caption{Example of experimental results for the use of PD control (cf. Equation \eqref{eq:PD_output}) for tracking dynamic walking on hardware. (left) Joint tracking for DURUS during multicontact walking \cite{reher2016durusmulticontact}. (right) Leg length output tracking on Cassie while walking with compliant HZD gait \cite{reher2019dynamic}. }
	\label{fig:time_trajectories}
\end{figure*}

\begin{summary}[Experimental Highlight: Trajectory Tracking]
The trajectories found in \secref{sec:nonlinear_planning} for HZD are well suited to feedback controllers for output tracking problems. To demonstrate the simplest and yet effective implementations of PD controllers which have been successful in realizing dynamic walking, we show experimental results on hardware for DURUS and Cassie in \figref{fig:time_trajectories}. 
This shows that for controllers in the joint space (\eqnref{eq:PD_joint}) and output space (\eqnref{eq:PD_output}) dynamic walking is can be achieved by simply tracking the designed motion. 
\end{summary}

\subsubsection{Inverse Dynamics} \label{sec:inverse_dynamics}

While PD control is sufficient for many applications, it fails to \textit{explicitly} consider the model of the robot and the constraints under which it operates. 
\textit{Inverse dynamics} is a widely used method to approach model-based controller design for achieving a variety of motions and force interactions, typically in the form of task-space objectives.  Given a target behavior, the dynamics of the robotic system are inverted to obtain the desired torques. In most formulations, the system dynamics are mapped onto a support-consistent manifold using methods such as the dynamically consistent support null-space \cite{sentis2007synthesis}, linear projection \cite{aghili2005unified}, and orthogonal projection \cite{mistry2010inverse}. When prescribing behaviors in terms of purely task space objectives, this is commonly referred to as task- or operational-space control (OSC) \cite{khatib1987unified}.  In recent work, variations of these approaches have been shown to allow for high-level tasks to be encoded with intuitive constraints and costs in optimization based controllers, some examples being \cite{apgar2018fast, kuindersma2016optimization, feng2015optimization, koolen2016design, herzog2016momentum}.

%
A benefit of inverse dynamics approaches to feedback control on robotic systems is that 
low gain 
feedback control can be used, while feedforward terms which respect the constrained rigid body dynamics of the physical system are used to produce the majority of the control action. 
%
If the walking is not significantly disturbed from the planned motion found in \secref{sec:nonlinear_planning}  then a linear null-space projection operator $P_F(q)$ can be used to eliminate the contact forces $\lambda$ from the floating-base dynamics in \eqnref{eq:eom_continuous} \cite{aghili2005unified},
using QR decomposition \cite{mistry2010inverse} to obtain an orthogonal projection into the null-space of $J_h(q)$.

The inverse dynamics problem can also be posed using a quadratic program (QP) to exploit the fact that the instantaneous dynamics and contact constraints can be expressed linearly with respect to a certain choice of decision variables. Specifically, let us consider the set of optimization variables $\mathcal{X} = [ \ddot{q}^T, u^T, \lambda^T]^T \in \Xext := \mathbb{R}^n \times U \times \mathbb{R}^{m_h}$, which are linear with respect to \eqnref{eq:eom_continuous} and \eqnref{eq:Jeom},
\begin{align}
    \begin{bmatrix} D(q) & -B & - J_h(q)^T  \\ J_h(q) & 0 & 0 \end{bmatrix} \mathcal{X} + \begin{bmatrix} H(q,\dot{q}) \\ \dot{J}_h(q)\dot{q} \end{bmatrix} = 0,  \label{eq:IDform}
\end{align}
and a positional objective in the task space of the robot written as $J_y(q) \ddot{q} + \dot{J}_y(q,\dot{q}) \dot{q} - \ddot{y}_2^* = 0$,
where $J_y(q) =  \partial y^a / \partial q$ and $\ddot{y}^* = K_P y + K_D \dot{y}$ is a PD control law which can be tuned to achieve convergence.
An additional benefit to using an optimization-based approach is the ability to include feasibility constraints such as the friction cone (\eqnref{eq:cone_friction}). However, this constraint is nonlinear, and cannot be implemented as a linear constraint. 
An alternative solution is to use a \textit{pyramidal friction cone} approximation \cite{grizzle2014models},
\begin{align}
    \mathcal{P} = \left\{ \left. ( \lambda_x, \lambda_y, \lambda_z )\in \mathbb{R}^3 \right| \lambda_z \geq 0; |\lambda_x|, |\lambda_y| \leq \frac{\mu}{\sqrt{2}} \lambda_z \right\}, \label{eq:pyramid_friction}
\end{align}
which is a more conservative model than the friction cone but is advantageous in that it is a linear inequality constraint. 
In it's most basic case, we can combine these elements to pose this QP tracking problem as:
\begin{align*}
    \mathcal{X}^{\ast}(x) = \argmin_{\mathcal{X}\in \Xext} \quad & || J_y(q) \ddot{q} + \dot{J}_y(q,\dot{q}) \dot{q} - \ddot{y}^* ||^2  + \sigma W(\mathcal{X}) \tag{\text{ID-QP}}\\
   \textrm{ s.t.} \quad & \text{\eqnref{eq:IDform}} \tag{\text{(System Dynamics)}} \\
   & u_{\mathrm{min}} \leq u \leq u_{\mathrm{max}} \tag{\text{(Torque Limits)}} \\
    & \text{\eqnref{eq:pyramid_friction}} \tag{\text{(Friction Pyramid)}}
\end{align*}
where $W(\mathcal{X})$ is included as a regularization term with a small weight $\sigma$ such that the problem is well posed. 
Although this kind of control satisfies the contact constraints of the system and yields an approximately optimal solution to tracking task-based objectives, it does not provide formal guarantees with respect to stability. In increasingly dynamic walking motions this becomes an important consideration,  
wherein impacts and footstrike can destabilize the system requiring more advanced nonlinear controllers.

\subsubsection{Control Lyapunov Functions for Zeroing Outputs}


The methods presented thus far demonstrate how feedback control can drive the dynamics of the robotic system to behave according to the planned motions found in \secref{sec:motion_planning}. 
However, these designs often intrinsically ignore the natural dynamics of the system, which is a critical component in the realization of efficient and dynamic walking. 
Thus, for practical systems, additional considerations for selecting our control input are often required. 
\textit{Rapidly exponentially stabilizing control Lyapunov functions} (RES-CLFs), were introduced as methods for achieving (rapidly) exponential stability for walking robots \cite{ames2014rapidly, ames2012control}.  A function, $V$, is a RES-CLF if it satisfies: 
\begin{eqnarray}
    && \underline{\gamma} \| x \|^2 \leq V_{\epsilon}(x) \leq \frac{\overline{\gamma}}{\epsilon^2} \| x \|^2 \\
  \inf_{u \in U}\Big[  \dot{V}_{\epsilon}(x,u)    \Big]    &= &  \inf_{u \in U} \Big[
    \underbrace{\frac{\partial V_{\epsilon}}{\partial x}(x) f(x)}_{L_f V_{\epsilon}(x)}  + 
    \underbrace{\frac{\partial V_{\epsilon}}{\partial x}(x) g(x)}_{L_g V_{\epsilon}(x)} u
    \Big]\leq - \frac{\gamma}{\epsilon} V_{\epsilon}(x) 
\end{eqnarray}
for $\underline{\gamma}, \overline{\gamma},\gamma > 0$, and $0 < \epsilon < 1$ a control gain that allows one to control the exponential convergence of the CLF, and is the basis for the term ``rapid'' in RES-CLF.  Importantly, if the robotic system is feedback linearizable per Section \ref{sec:hzd} it automatically yields a Lyapunov function.  In particular, defining $\eta(x) := (y(x)^T, \dot{y}(x)^T)^T$, we obtain RES-CLF: $V_{\epsilon}(x) = \eta(x)^T P_{\epsilon} \eta(x)$ where $P_{\epsilon} = \mathbf{I}_{\epsilon} P \mathbf{I}_{\epsilon}$ with $\mathbf{I}_\epsilon := \mathrm{diag}\left( \frac{1}{\epsilon} \mathbf{I}, \mathbf{I} \right)$ and $P$ the solution to the continuous time algebraic Riccati equations (CARE) for the linear system $\ddot{y} = \mu$ obtained by feedback linearization in \eqnref{eq:FBL}.

The advantage to controller synthesis with CLFs is that they yield an entire class of controllers that provably stabilize periodic orbits for hybrid system models of walking robots, and can be realized in a pointwise optimal fashion via optimization based controllers.  In particular, consider the set of control inputs:
\begin{align}
    K_{\epsilon}(x) = \{u \in U : L_f V_{\epsilon}(x) + L_g V_{\epsilon}(x) u \leq - \frac{\gamma}{\epsilon} V_{\epsilon}(x)  \}, \label{eq:ES_clf_u_class}
\end{align}
which is a set of stabilizing controllers.  To see this, note that for $u^*(x) \in K_{\epsilon}(x)$:
\begin{align}
\label{eq:Vconvergence}
\dot{V}_{\epsilon}(x,u^*(x)) \leq - \frac{\gamma}{\epsilon} V_{\epsilon}(x)
\qquad & \Rightarrow \qquad V(x(t)) \leq e^{-\frac{\lambda}{\epsilon} t} V(x(0)) \\
\qquad & \Rightarrow \qquad 
\| \eta(x(t)) \| \leq \frac{1}{\epsilon} \sqrt{ \frac{\lambda_{\max}(P)}{\lambda_{\min}(P)}} e^{- \frac{\gamma}{2 \epsilon} t } \| \eta(0) \|. 
\nonumber
\end{align}
Thus, this gives the set of control values that exponentially stabilize the outputs and we can control the convergence rate via $\epsilon$.  The selection of an appropriate choice for the ``best'' control value possible leads to the notion of optimization based control with CLFs.

The advantage of \eqnref{eq:ES_clf_u_class} is that it gives a set of controllers that result in stable walking on bipedal robots. 
That is, for any $u \in K(x)$ the hybrid system model of the walking robot, per the HZD framework introduced in \ref{sec:hzd}, has a stable periodic gait given a stable periodic orbit in the zero dynamics \cite{ames2014rapidly}. This suggests an optimization-based framework nonlinear controller synthesis, with specific application to dynamic locomotion.  Specifically, the optimization formulation of CLFs allows for additional constraints and objectives to be applied 
as a QP with the form (as first introduced in \cite{ames2013towards}):
\begin{align*}
    u^{\ast} = \argmin_{u\in U \subset \mathbb{R}^m} \quad & u^T H(x) u + \rho \delta^2 \tag{CLF-QP}\\
    \mathrm{s.t.} \quad & L_f V_{\epsilon}(x) + L_g V_{\epsilon}(x)u \leq - \frac{\gamma}{\epsilon} V_{\epsilon}(x) + \delta \tag{\text{(CLF Convergence)}}\\
                        & u_{\mathrm{min}} \leq u \leq u_{\mathrm{max}} \tag{\text{(Torque Limits)}} \\
                        & \text{\eqnref{eq:pyramid_friction}} \tag{\text{(Friction Pyramid)}}
\end{align*}
where $H(x)$ is a user specified positive-definite cost, $\delta$ is a relaxation to the convergence constraint which can be added if infeasibility of the solution is a concern, and $\rho>0$ is a large value that penalizes violations of the CLF constraint. If the relaxation term is included then the formal guarantees on convergence are no longer satisfied in lieu of achieving pointwise optimal control actions which satisfy the physical constraints of the robot. Ground reaction forces on the robot also
appear in an affine fashion in the dynamics; thus one can also use the CLF-based
QP framework in the context of force control \cite{ames2013towards}.

\begin{figure*}[b]
	\centering
	\includegraphics[width= 1.0 \columnwidth]{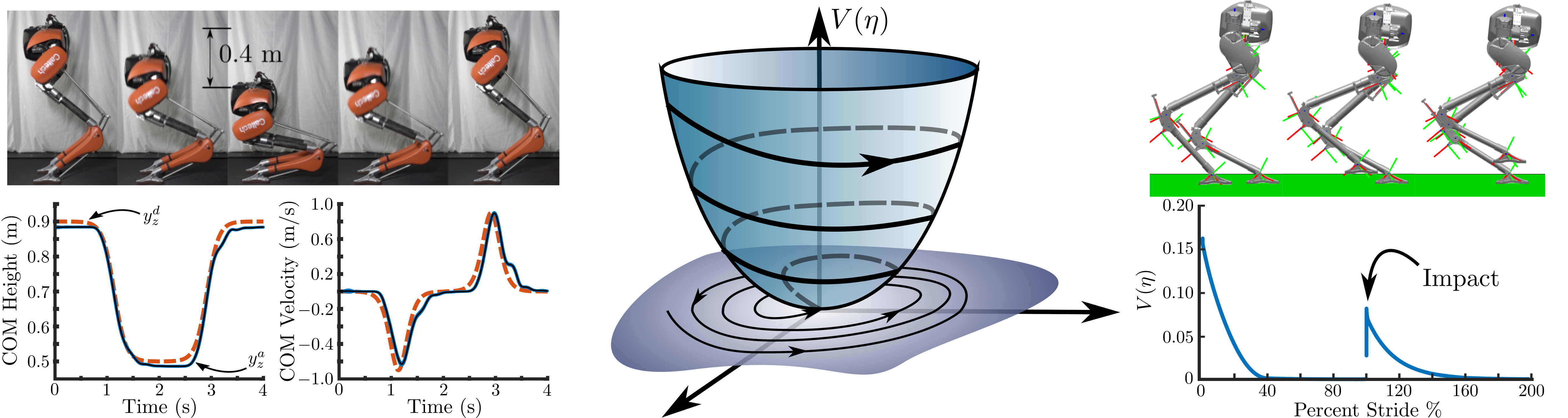}
	\caption{ A visualization of a CLF driving a Lyapunov function to zero, with data from an experimental implementation (left) and walking simulation (right). Because CLFs consider the model to enforce convergence, outputs are closely tracked with minimal error (left). The rapid exponential zeroing the outputs (\eqnref{eq:Vconvergence}) is critical to achieve sufficient convergence before impact (right).  }
	\label{fig:cassie_clf}
\end{figure*}

The CLF-based controllers presented throughout this section have recently been explored of interest for application on hardware, because much like the optimization controllers of \secref{sec:inverse_dynamics} they can be solved in real-time. Experimental results have been shown on MABEL \cite{ames2014rapidly,galloway2015torque} and DURUS-2D \cite{cousineau2015realizing}, with recent results indicating how robust formulations can be used \cite{nguyen2020optimal} and how alternative representations can make the problem more tractable for implementation on 3D robots \cite{reher2019inverse}. Additionally, it was implemented at over $5$ kHz as an embedded level controller on series elastic actuators \cite{ames2014quadratic},  indicating possible future uses on explicitly controlling compliant dynamic walking. 
The use of CLFs has also been used to automatically generate stable walking gaits through SLIP approximations \cite{hereid2014embedding}, and also to enforce planned motions for reduced order models \cite{xiong2018coupling} along with realizing 3D bipedal jumping experimentally on Cassie \cite{xiong2018bipedal}.

\begin{summary}[Experimental Highlight: Real-Time CLF QP Control]
We highlight the application of CLF based QPs on hardware in real-time in the context of dynamic crouching maneuvers on Cassie, shown in \figref{fig:cassie_clf} \cite{reher2019dynamic}. 
Because the CLF-QP can be run at a sufficient control frequency (in this case at $1$ kHz), these experiments show how convergence properties combined with inclusion of the model can lead to desirable tracking performance on complex bipedal robots.
These methods are directly extensible to tracking walking trajectories, where the constrained pointwise optimization can select torques which satisfy the contact constraints 
governed by the discrete structure of the hybrid system model 
(see \figref{fig:periodic_and_hybrid}).
\end{summary}

\subsection{Stabilizing Walking with Trajectory Modification} \label{sec:footplacement}

\begin{figure*}[b]
	\centering
	\includegraphics[width= 1.0 \columnwidth]{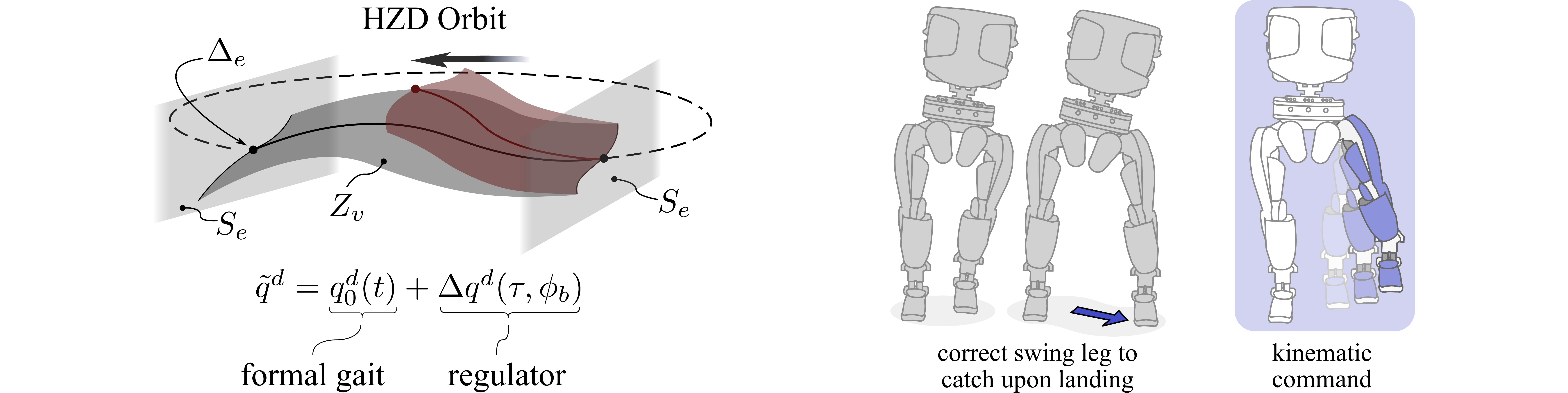}
	\caption{A visualization of how a regulator action is used to drive a perturbed zero dynamics surface back to the nominal motion (left). This can take the form of direct joint changes or Cartesian foot placement (right), making a kinematic adjustment in response to torso lean or velocity. }
	\label{fig:regulation_concept}
\end{figure*}

The prior sections detailed how dynamic walking behaviors are formulated, synthesized and tracked; yet, these components alone are often not sufficient to realize sustained and robust robotic walking on hardware.  The final step in achieving robustness involves the ``artful implementation'' of modifying the desired behavior to account for unknown and unmodified disturbances -- both specific to the hardware, e.g., unmodified compliance, and in the external environment, e.g., rough terrain. 
Approaches such as MPC planners and analytical expressions for the CP, presented in \secref{sec:step_plan_linear}, can be evaluated in real-time to adapt the motion of the robot to avoid falling or recover from large pushes \cite{stephens2010push}. In these cases the planning and the real-time compensation are inherently tied \cite{wieber2006online}, though they still are re-planning over an approximate the model of the robot and can lead to constrained motions which are prohibitive to truly dynamic walking. 
%
On the other hand, while the nominal trajectories of offline plans which consider the full-body continuous and hybrid dynamics are generated with high fidelity models (such as the motions found via \secref{sec:nonlinear_planning})
it is 
evident in experimental trials that some 
additional feedback is crucial to stabilizing the robot for sustained periods of walking. 
These nominal trajectories are often superimposed with some form of \textit{regulator} in order to overcome uncertainties due to model mismatch and tracking errors, typically in the form of adding trajectory level feedback (see Fig. \ref{fig:regulation_concept}). 
The development of these regulators is a largely heuristic task, but has often proven critical to stability on hardware. 
A variety of different regulators have proven useful, though 
implementation largely dependents on the robotic system and 
desired behavior.

When performing dynamic maneuvers it is inevitable that the actual linkages of humanoid robots, which have large masses and inertias, can subject rotational joints to backlash and unsensed compliance. For these problems, using an experimentally measured stiffness coefficient to augment commanded positions based on anticipated torque at the joint has shown to be an effective compensation strategy \cite{reher2016durusmulticontact,johnson2015team}. 
The combination of uncertainty in the kinematics and dynamics of the robot can also lead to predictable issues with gait timing on periodic walking behaviors. 
For dynamic walking which has been planned with a monotonic phase variable $\tau(q)$ dependent on the state of the robot, there can be a large amount of uncertainty with regards to the estimation of the floating-base coordinates and therefore the phase \cite{ma2017bipedal,kolathaya2016time}. In these cases it can be beneficial to employ a combination of time and state based progression of the variable \cite{da20162d}.

\begin{figure*}[t]
	\centering
	\includegraphics[width= 1.0 \columnwidth]{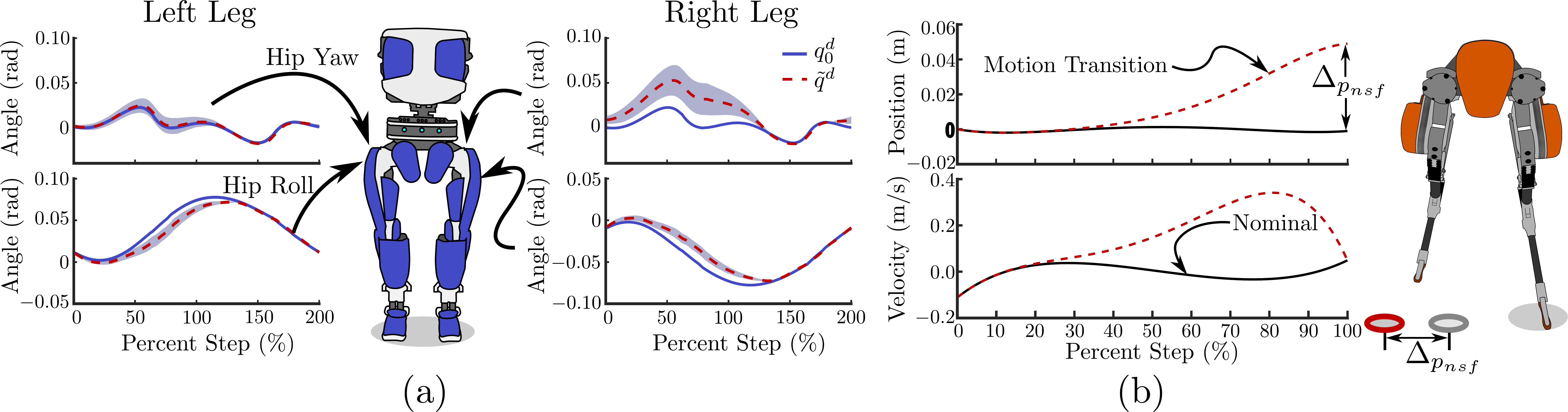}
	\caption{
	(a) Experimental data from walking on the DURUS humanoid \cite{reher2016durusrealizing}, where the shaded region is $1$ std. dev. of over 200 steps. Trajectories are modified by a regulator proportional to torso lean. (b) An example of a motion transition \cite{powell2013speed} applied to an output for the swing foot position. }
	\label{fig:motion_transitions}
\end{figure*}

Another type of regulation comes in form of small modifications to the shape of the robot (i.e. superimposed perturbations to virtual constraints) from stride-to-stride. 
How this can be conceptually interpreted within the HZD framework is shown in \figref{fig:regulation_concept}, where control designers seek to shape a perturbed zero dynamics surface such that the hybrid system returns to an orbit which satisfies hybrid invariance. 
In early developments for control of HZD walking, the restricted Poincar\`e map was viewed as a discrete-time control system \cite{chevallereau2005asymptotically}. Through consideration of the linearized map at the fixed point $x^*$ (see \eqnref{eq:poincare_linearized}), a discrete LQR algorithm can then be used to acquire a feedback gain to modify the configuration of the next footstrike \cite{ramezani2014performance}. This can be straightforward to design for 2D robots, but extensions to 3D become more difficult. 
Perhaps the most common approach is to instead utilize foot placement routines inspired by Raibert \cite{raibert1984experiments}. 
This simple deadbeat step-to-step controller most often takes the form of a discrete PD controller to augment the footstrike locations in the sagittal and frontal planes during locomotion: 
\begin{align}
    \Delta p_{nsf} &= \tilde{K}_p (\bar{v}_k - v_{\mathrm{ref}} ) + \tilde{K}_d (\bar{v}_k - \bar{v}_{k-1}),
\end{align}
where the average velocity of the current step $\bar{v}_k$ and previous step $\bar{v}_{k-1}$ are computed directly from an estimate of the floating-base velocity, and the reference velocity $v_{\mathrm{ref}}$ is taken from the nominal trajectory. 
In addition, because outputs for HZD walking are typically parameterized by a B\'ezier polynomial, the update value $\Delta p_{nsf}$ can directly augment the last two parameters of the corresponding output polynomials \cite{powell2013speed}. This kind of smooth transition is demonstrated in \figref{fig:motion_transitions}(b), where the position has been smoothly modified, but the velocity at impact will remain the same. 
This simple foot placement regulator has been successfully implemented on several dynamic walking robots \cite{da20162d, Rezazadeh2015, reher2019dynamic}. 
Rather than considering hand-tuned regulation, the notion of \textit{nonholonomic virtual constraints} was introduced \cite{griffin2015nonholonomic}, aiming to formalize a representation of virtual constraints which are insensitive to a predetermined and finite set of terrain variations and velocity perturbations.  Implementation of this approach required intensive optimizations, as the walking was made to be stable amid a variety of perturbations in each step of the optimization. 
This type of output-level feedback has also been proven to be successful in a more directly hand-tuned fashion, such as in \cite{reher2016durusrealizing,reher2016durusmulticontact}, where the position-level feedback of the outputs was governed by a proportional gain with respect to the pitch and roll of the robot's torso. The superimposed motion will then be zero if the walking is directly on the orbit, but will smoothly apply a superimposed positional command if necessary. 
One interpretation of this regulator feedback is simply that $y_a$ has been made a function of the floating base coordinates of the robot, with an example shown on the DURUS humanoid in \figref{fig:motion_transitions}(a).


\begin{figure*}[t!]
	\centering
	\includegraphics[width= 1.0 \columnwidth]{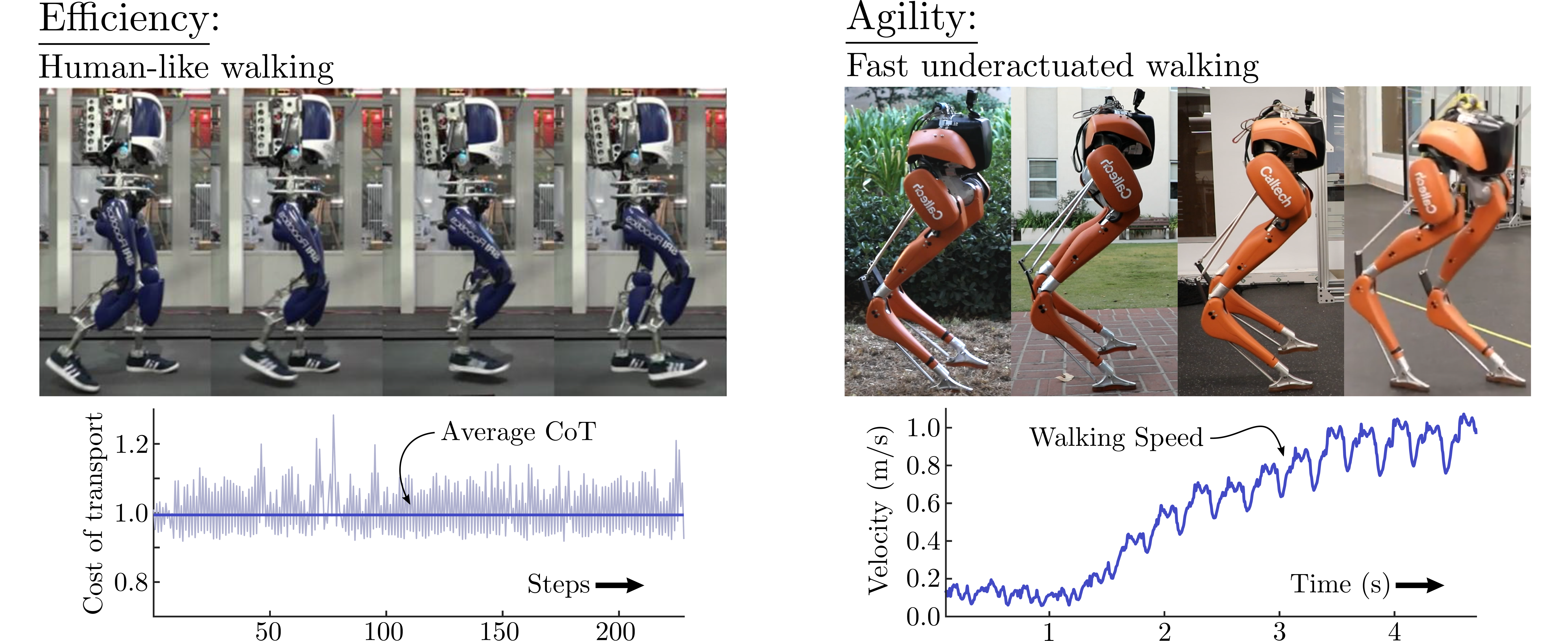}
	\caption{Experimental examples of dynamic walking on DURUS and Cassie. Gait tiles are provided, showing the robots in various phases of their natural strides along with plots of data detailing the efficiency of the walking for DURUS and a plot of the sagittal walking velocity on Cassie. }
	\label{fig:dynamic_walking_example}
\end{figure*}

\begin{summary}[Experimental Highlight: Dynamic Walking]
The experimental highlights considered throughout this this paper culminates with efficient and agile locomotion on DURUS and Cassie, as shown in \figref{fig:dynamic_walking_example}.
The multicontact walking on DURUS demonstrated efficiency as evidenced by an exceptionally low cost of transport \cite{reher2016durusmulticontact}, achieved by leveraging hybrid models, to closed loop optimization, to real-time feedback controllers and regulators. 
The compliant walking on Cassie \cite{reher2019dynamic} demonstrates agility, through a wide range of walking speeds up to $1$ m/s, along with the ability to walk on unplanned rough terrain outdoors. 
\end{summary}

\section{CONCLUDING REMARKS} \label{sec:discussion}

This review article outlined the general methodology for achieving dynamic walking on bipedal robots.  As outlined in ``Summary Points,'' we begin by considering reduced order models that capture the essentials of locomotion -- yet these models are not sufficient for handling the full complexity of walking robots.  This led to full order models that include impacts as represented by hybrid systems, wherein we considered hybrid zero dynamics.  To generate walking gaits with these models, and corresponding dynamically feasible trajectories, the role of optimization was discussed.  Finally, connecting models with walking gaits, real-time controllers that enable hardware realization were discussed; 
ranging from 
simple control methods, 
to advanced QP based controllers, 
together with the modification of these nominal desired values due to uncertainty in the system and environment.  This end-to-end process was illustrated throughout on 
the bipedal robots DURUS and Cassie, 
wherein the translation to hardware and corresponding experimental results were highlighted.  

Looking forward, the process of realizing dynamic walking that is efficient and agile is ripe with opportunities.  Some of these challenges are highlighted in ``Future Issues.''  In essence, these can be subdivided into two categories: theoretic and practical.  The overarching goal, theoretically, 
is to formally and holistically extend the methodologies presented.  The hope is to, as a result, develop a framework that is capable of realizing aperodic dynamic motions that are stable and safe which are planned in real-time and robust to uncertainties in the robot and environment. 
From a practical perspective, hardware is ever improving and becoming more accessible.  This gives the ability to better test approaches for agile and efficient walking in real-world scenarios.  The goal is to finally realize the promise of dynamic walking: imbuing legged robots with the locomotion capabilities that will enable them to do everything from traversing everyday environments to exploring the cosmos. 

\begin{summary}[SUMMARY POINTS]
\begin{enumerate}
\item \emph{Reduced order models.} At the core of dynamic walking is the idea of reduced order models.  These are either hierarchical---representing desired behavior on simple models, e.g., inverted pendula and compass gait bipeds---or formally determined---low-dimensional systems rendered invariant by controllers, e.g., HZD. 
\item \emph{Full order nonlinear dynamics.} Bipedal robots are inherently nonlinear with hybrid dynamical behaviors.  These full order dynamics must be accounted for, either through assumptions thereon that yield reduced order models, through nonlinear controllers, or via optimization algorithms. 
\item \emph{Optimization for gait generation.} Reduced order models must be instantiated on the full order dynamics via optimization algorithms.  This can leverage reduced order models, exploit the full order dynamics, or any combination thereof.  Algorithms that allow for these optimization problems to be solved efficiently are essential in instantiating walking gaits on hardware platforms.  
\item \emph{Control laws for hardware realization.} Control laws allow for the generated gaits to, ultimately, be realized on hardware.  These can range from simple control laws to complex nonlinear real-time optimization-based controllers, and can be modulated via inspiration from reduced order models.  These control algorithms are the final step in realizing dynamic walking on bipedal robots. 
\end{enumerate}
\end{summary}

\begin{issues}[FUTURE ISSUES]
\begin{enumerate}
\item \emph{Generalized notions of stability and safety.}  The walking considered herein, and the notions of stability, was largely periodic in nature.  To better represent a wide variety of behaviors, the idea of stability should be extended to include aperiodic walking motions \cite{nguyen20163d,ames2017first}.  More generally, safety as represented by set invariance \cite{ames2016control} could provide a powerful tool for
more generally understanding locomotion. 
\item \emph{Real-time optimal gait planning.} It was seen that nonlinear constraint optimization plays an essential role in generate dynamic walking behaviors that leverage the full-body dynamics.  These methods have become very efficient, even allowing for online calculation in simple scenarios \cite{hereid2016online}.  Further improving computational efficiency will enable real-time implementation yielding new paradigms for gait generation. 
\item \emph{Bridging the gap between theory and practice.}  As indicated by the methods discussed in Section \ref{sec:regulation}, there is often an ``artful implementation'' step that translates model-based controllers to a form that can actually implemented on hardware.  Ideally, methods can be developed that allow the exact transcription of model-based methods to hardware in a robust fashion and without heuristics. 
\item \emph{Robustness, adaptation and learning.} 
Dynamic walking behaviors often work in isolated instances and predefined environments.  Translating these ideas to the real-world will require robustness to uncertainty -- both in the internal dynamics and external environment.  Adaptive and learning-based controllers can help mitigate model uncertainty and unplanned interactions with the world, from uncertain contact conditions to walking on surfaces with complex interactions, e.g., sand.
\item \emph{Real-world deployment of bipedal robots.} The ultimate challenge is the ability to deploy bipedal robots in real-world scenarios.  This ranges from everyday activities, to aiding humans, to venturing into dangerous environments.  Examples include bipedal robotics in a healthcare setting, e.g., exoskeletons for restoring mobility \cite{gurriet2018towards}, to humanoid robots capable of exploring Mars.  
\end{enumerate}
\end{issues}

\section*{DISCLOSURE STATEMENT}
The authors are not aware of any affiliations, memberships, funding, or financial holdings that might be perceived as affecting the objectivity of this review. 

\section*{ACKNOWLEDGMENTS}
The authors would like to thank the members of AMBER Lab whom have contributed to the understanding of robotic walking summarized in this paper.  Of special note are Shishir Kolathaya, Wenlong Ma, Ayonga Heried, Eric Ambrose and Matthew Powell for their work on AMBER 1, 2 and 3M and DURUS, from developing the theory, to computational methods, to experimental realization. 
Outside of AMBER Lab, the authors would like to thank their many collaborators.  Of particular note is Jessy Grizzle and the joint efforts on HZD and CLFs.
This work was supported over the years by the National Science Foundation, including awards: CPS-1239055, CNS-0953823, NRI-1526519, CNS-1136104, CPS-1544857.  Other support includes projects from NASA, DARPA, SRI, and Disney. 

\newpage

\bibliographystyle{style/ar-style3.bst}
\bibliography{main.bib}

\end{document}